%% file: main.tex
\pdfoutput=1

\documentclass[conference]{IEEEtran}
\IEEEoverridecommandlockouts

\usepackage[numbers,sort]{natbib}
\usepackage{amsmath,amssymb,amsfonts}
\usepackage{algorithmic}
\usepackage{graphicx}
\usepackage{textcomp}
\usepackage{xcolor}

\usepackage{booktabs}
\usepackage{bm} 
\usepackage{dsbda} 

\usepackage{xcolor} 
\newcommand{\Lhard}{\Ls_\text{hard}}
\newcommand{\Lsoft}{\Ls_\text{soft}} 

\newcommand{\cmow}{{(\mathrm{CMOW})}}
\newcommand{\cbow}{{(\mathrm{CBOW})}}
\newcommand{\hybrid}{{(\mathrm{Hybrid})}}
\newcommand{\bidicmow}{{(\mathrm{Bidi.\,CMOW})}}
\newcommand{\bidicbow}{{(\mathrm{Bidi.\,CBOW})}}
\newcommand{\bidihybrid}{{(\mathrm{Bidi.\,Hybrid})}}
\newcommand{\fwd}{{(\mathrm{fw})}}
\newcommand{\bwd}{{(\mathrm{bw})}}
\newcommand{\cat}{\mid\mid}

\newcommand{\CMOWCBOWHybrid}{CMOW/CBOW-Hybrid\xspace}

\usepackage{hyperref}
\usepackage{url}

\usepackage{xcolor,colortbl}

\usepackage{times}
\usepackage{latexsym}

\usepackage[T1]{fontenc}

\usepackage[utf8]{inputenc}

\usepackage{microtype}

\title{
General Cross-Architecture Distillation of Pretrained Language Models into Matrix Embeddings
}

\author{\IEEEauthorblockN{Lukas Galke}
\IEEEauthorblockA{
\textit{Max Planck Institute for Psycholinguistics} \\ Nijmegen, Netherlands\\
lukas.galke@mpi.nl}
\and
\IEEEauthorblockN{Isabelle Cuber, Christoph Meyer, Henrik Ferdinand N\"olscher, \\ Angelina Sonderecker, Ansgar Scherp}
\IEEEauthorblockA{ \textit{University of Ulm}, 
Germany \\
\{isabelle.cuber,christoph-1.meyer,henrik-1.noelscher,angelina.sonderecker,\\ ansgar.scherp\}@uni-ulm.de}

}

\begin{document}
\maketitle

\begin{abstract}
Large pretrained language models (PreLMs) are revolutionizing natural language processing across all benchmarks. However, their sheer size is prohibitive for small laboratories or for deployment on mobile devices. Approaches like pruning and distillation reduce the model size but typically retain the same model architecture. In contrast, we explore distilling PreLMs into a different, more efficient architecture, Continual Multiplication of Words (CMOW), which embeds each word as a matrix and uses matrix multiplication to encode sequences. We extend the CMOW architecture and its \CMOWCBOWHybrid variant with a bidirectional component for more expressive power, per-token representations for a general (task-agnostic) distillation during pretraining, and a two-sequence encoding scheme that facilitates downstream tasks on sentence pairs, such as sentence similarity and natural language inference. Our matrix-based bidirectional \CMOWCBOWHybrid model is competitive to DistilBERT on question similarity and recognizing textual entailment, but uses only half of the number of parameters and is three times faster in terms of inference speed. We match or exceed the scores of ELMo for all tasks of the GLUE benchmark except for the sentiment analysis task SST-2 and the linguistic acceptability task CoLA. However, compared to previous cross-architecture distillation approaches, we demonstrate a doubling of the scores on detecting linguistic acceptability. This shows that matrix-based embeddings can be used to distill large PreLM into competitive models and motivates further research in this direction.
\end{abstract}

\section{Introduction}
Large pretrained language models~\cite{bert,T5} (PreLMs) have emerged as de-facto standard methods for natural language processing~\cite{glue,superglue}.
The common strategy is to pretrain models on enormous amounts of unlabeled text before fine-tuning them for downstream tasks.
However, the drawback of PreLMs is that the models are becoming larger and larger with up to several billion parameters~\cite{GPT-3}.
This comes with high environmental and economic costs~\cite{DBLP:conf/acl/StrubellGM19} and puts development and research in the hands of a few global players\extended{~with rich resources} only \cite[pp. 10-12]{foundation-models}.
Even though a single pretrained model can be reused for multiple downstream tasks, the sheer model size is often prohibitive.
The immense resource requirements prevent the use of these models in small-scale laboratories and on mobile devices, which is tied to privacy concerns~\cite{DBLP:conf/nips/Sanh0R20}.

There is a need for more efficient models or compressed versions of large models to make AI research more inclusive and energy-friendly while fostering deployment in applications.
Reducing the size of PreLMs using knowledge distillation~\cite{knowledgedistillation} or model compression~\cite{modelcompression} is an active area of research~\cite{distilbert,tinybert,sun2020mobilebert}. 
It is reported that companies, such as Google, use distillation of PreLM to deploy their large models for productive use, \ie for services that have strong requirements in terms of low latency.\footnote{J. Devlin, ``Contextual Word Representations with
BERT and Other Pre-trained Language
Models'',
\url{https://web.stanford.edu/class/cs224n/slides/Jacob_Devlin_BERT.pdf}}
Both knowledge distillation and model compression can be described as teacher-student setups~\cite{knowledgedistillation,modelcompression}.
The student is trained to imitate the predictions of the teacher while using less resources.
Typically, a large PreLM takes the role of the teacher while the student is a smaller version of the same architecture.
Sharing the same architecture between the student and the teacher enables the use of dedicated distillation techniques, \eg aligning the representations of intermediate layers~\cite{distilbert,sun2020mobilebert}.

However, using more efficient architectures as student has already shown promising results, such as the task-specific distillation approaches by Tang~\etal~\cite{tang2019distilling} and Wasserblatt~\etal~\cite{boundariesbertdistillation}.
In their works, the student models are LSTMs~\cite{DBLP:journals/neco/HochreiterS97} or models based on a continuous bag-of-words representation (CBOW)~\cite{DBLP:conf/icml/CollobertW08,word2vec}.
On the one hand, LSTMs are difficult to parallelize as they need at least $\mathcal{O}(n)$ sequential steps to encode a sequence of length $n$.
On the other hand, CBOW-based models are \textit{not order-aware}, \ie cannot distinguish sentences with the same words but in different order (``cat eats mouse'' vs. ``mouse eats cat'' are treated equivalent).
There are, however, efficient models such as Mai~\etal's continual multiplication of words (CMOW) that \textit{do capture word order} by representing each token as a matrix~\cite{cmow}, instead of a vector as in CBOW. 
A sequence in CMOW is modeled by the non-commutative matrix multiplication~\cite{DBLP:conf/acl/RudolphG10}, which makes the encoding of a sequence dependent on the word order. We denote such models as \emph{matrix embeddings}.

We extend Mai et al.'s work and investigate how order-aware matrix embeddings can be used as student models in \textit{cross-architecture distillation} from large PreLM teachers.
This complements the existing body of works that focused predominantly on \textit{same-architecture distillation} (see discussion in Section~\ref{sub:kd}).
Furthermore, all previous cross-architecture distillation approaches are task-specific, whereas we also explore general distillation.
We aim to understand to what extent order-aware embeddings are suitable to capture the teacher signal of a large PreLM such as BERT~\cite{bert}.
To this end, we extend Mai~\etal's \CMOWCBOWHybrid model~\cite{cmow}, which is a hybrid variant unifying the strength of CBOW and CMOW, with a bidirectional representation of the sequences.
Furthermore, we add the ability to emit per-token representations to facilitate the use of a modern masked language model objective~\cite{bert}.

We investigate both task-agnostic \emph{general distillation}, \ie the distillation is applied during pretraining on unlabeled text, and \emph{task-specific distillation}, when an already fine-tuned PreLM is distilled per task.
We further introduce a two-sentence encoding scheme to CMOW so that it can deal with sentence similarity and natural language inference tasks.

Our results show that large PreLMs can be distilled into efficient order-sensitive embedding models, achieving a performance that is competitive to ELMo~\cite{DBLP:conf/naacl/PetersNIGCLZ18} on the GLUE benchmark.
On the QQP and RTE tasks, embedding-based models even challenge other size-reduced BERT models such as DistillBERT.
In summary, our contributions are:
\begin{itemize}
  \setlength\itemsep{0em}
    \item We extend order-aware embedding models with bidirection and make them amenable for masked language model pretraining. 
     \item We explore using order-aware embedding models as student models in a cross-architecture distillation setup with BERT as a teacher and compare general and task-specific distillation.
    \item We introduce the first encoding scheme that enables \CMOWCBOWHybrid to deal with two-sentence tasks (20\% increase over the naive approach).
    \item Our results show that the best distilled embedding models are on-par with more expensive models such as ELMo.
    We outperform DistilBERT on the QQP and RTE tasks of the GLUE benchmark, while having a much higher encoding speed (thrice as high as DistilBERT).
\end{itemize}

Below, we introduce our embedding models, our cross-architecture distillation setup, and our two-sequence encoding scheme.
The experimental procedure is described in Section~\ref{sec:apparatus}.
The results are reported in Section~\ref{sec:results} and discussed in Section~\ref{sec:discussion}, where we also relate our work to the literature.

\section{Problem Formulation}\label{sec:cad:problem}
We study the problems of transfer learning and knowledge distillation, which we briefly describe below.

\subsection{Transfer Learning: Pretraining and Fine-Tuning} In a pretraining stage, a language model is trained on large amounts of unlabeled text in a self-supervised manner, \eg by predicting left-out words.
The pre-trained model is then transferred to downstream tasks. The parameters of the transferred model are fine-tuned to the respective task.
For each task, a fresh copy of the pretrained model is used as a starting point.

\begin{itemize}
  \item Step 1: Pretraining. Train model $f_\theta$ using unlabeled text.
  \item Step 2: Fine-tuning. For each downstream task $\gT$, continue training the model $f_\theta$ from pretraining.
  The model $f_\theta$ is extended by a task-specific classification head, which is initialized randomly.
\end{itemize}

The task $\gT$ can be an arbitrary supervised downstream tasks, where paired training data is available.
In particular, that includes two-sequence tasks such as natural language inference or textual similarity.
The performance measure depends on the respective task. We will use the tasks from the GLUE benchmark~\cite{glue} to evaluate our models.

\subsection{Knowledge Distillation}\label{sub:kd}
The problem of knowledge distillation~\cite{knowledgedistillation} or model compression~\cite{modelcompression} refers to learning a smaller model $g$ that imitates the behavior of a larger model $f$, such that desirably $g(\vx) \approx f(\vx)$.
The smaller model $g$ is called the student and the larger model $f$ is called the teacher.

The distillation is carried out by aligning the teacher's and student's output, \eg via a loss term $\Ls(f(\vx), g(\vx))$, which we call the teacher signal.
There are more techniques to foster distillation such as using the teacher's weights as initialization for the student~\cite{distilbert,tinybert,sun2020mobilebert}, but those are only applicable when teacher and student are of the same model architecture. When student and teacher models have a different architecture, we call that cross-architecture distillation.

To contextualize knowledge distillation with transfer learning,
we adopt the distinction between general distillation and task-specific distillation from Tang~\etal~\cite{tang2019distilling}.

\begin{itemize}
  \item General distillation: 
  The distillation is carried out \emph{only} in the (self-supervised) pretraining stage.  After pretraining, the teacher is not needed anymore. The student can be fine-tuned to downstream tasks independently.
  \item Task-specific distillation: 
  In task-specific distillation, the teacher model can be consulted during fine-tuning for each of the  downstream tasks. The respective task's training data are used for (supervised) distillation.
\end{itemize}

Note that the model for task-specific distillation can still be initialized with a model obtained by general distillation. However, this also falls under task-specific distillation, because, after all, the teacher has to be consulted during fine-tuning. 

Distinguishing between general and task-specific distillation is important because this affects how the methods can be applied in practice.
Imagine that we want to fine-tune a model for a downstream task on a mobile device.
A model from general distillation would be able to learn new tasks on its own.
With a task-specific distillation approach, the larger teacher model would need to be consulted on the mobile device.
Thus, both approaches differ in how they can be applied in practice and should be considered separately.

\section{Methods}
\label{sec:methods}

First, we introduce our bidirectional extension to the 
\CMOWCBOWHybrid model.
Subsequently, we introduce our approach for cross-architecture distillation that we use during the pretraining and fine-tuning stages. 
Finally, we introduce a two-sentence encoding scheme for order-aware embedding models that is crucial for fine-tuning on downstream tasks with paired sentences.

\subsection{Extending the Order-Aware Embedding Models}\label{sub:cmow-hybrid-bidi}
We extend the \CMOWCBOWHybrid embeddings of Mai~\etal~\cite{cmow} with bidirection and the ability to emit per-token representations as a foundation for cross-architecture distillation.
\CMOWCBOWHybrid embeddings, our baseline model, are a combination of matrix embeddings and vector embeddings.
Compared to vector-only embeddings, the word order can be captured because matrix multiplication is non-commutative.
Given a sequence $s$ of $n$ tokens with each token $s_j$ having its corresponding matrix-space embedding $\mX_j \in \sR^{d\times d}$ and vector-space embedding $\vx_j \in \sR^{d_\mathrm{vec}}$,
the \CMOWCBOWHybrid embedding of a sequence of length $n$ is the multiplication of embedding matrices $\mX_i$ concatenated (symbol $\cdot || \cdot$) to the sum of embedding vectors $\vx_i$:
\begin{align*}
    \mH^\cmow &:= \mX_1^\cmow \cdot \mX_2^\cmow \cdots \mX_n^\cmow\\
    \vh^\cbow &:= \sum_{1\leq j\leq n} \vx_j^\cbow\\
    \vh^\hybrid &:= \operatorname{flatten}\left(\mH^\cmow \right) \cat{} \vh^\cbow 
\end{align*}
where $\operatorname{flatten}$ collapses the matrix into a vector.
The original work on CMOW~\cite{cmow} has extensively analyzed the CMOW and CBOW components individually and found that joint training is generally preferable. The CMOW and CBOW components can have different dimensionalities because they are combined by concatenation. We initialize each matrix $\mX_j$ as identity plus Gaussian noise $\mI_d + \mathcal{N}(0, \sigma_\mathrm{init}^2)$ with $\sigma_\mathrm{init} = 0.01$.

\begin{figure*}[htpb]
  \centering
  \includegraphics[width=0.9\linewidth]{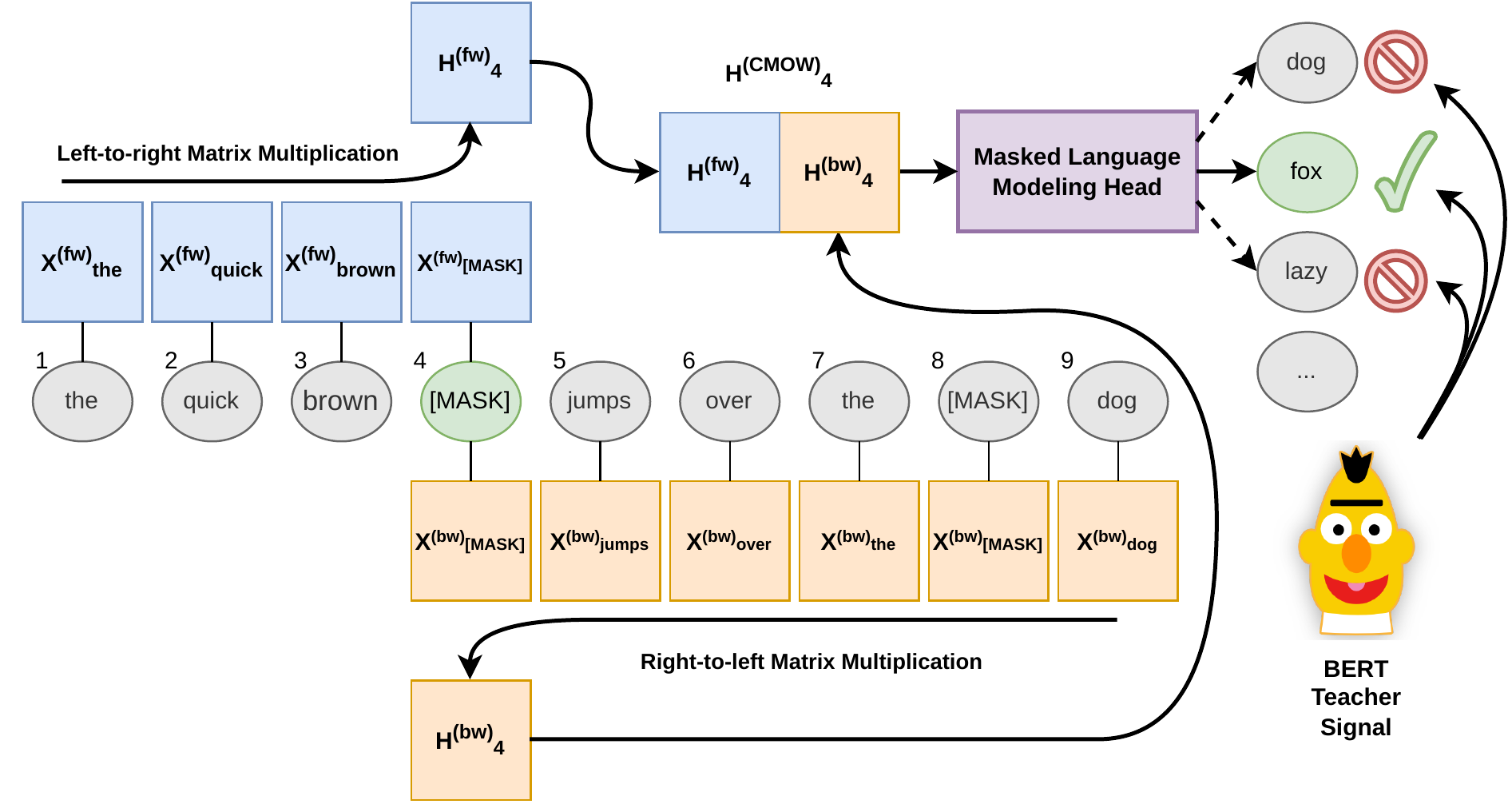}
  \caption{The bidirectional CMOW component of our proposed architecture during pretraining. In this example, the model predicts the masked token at position~4 ([MASK]) by concatenating forward and backward matrix embeddings, which are then fed into a masked language modeling head.}\label{fig:w2m:bidirection}
\end{figure*}

\paragraph{Proposed Model: Bidirectional \CMOWCBOWHybrid}
Inspired by the success of bidirectionality in RNNs~\cite{DBLP:journals/tsp/SchusterP97}, LSTMs~\cite{DBLP:conf/naacl/PetersNIGCLZ18}, and Transformers~\cite{bert}, we extend CMOW by a bidirectional component. 
Hence, we introduce a second set of matrix-space embeddings that are multiplied in reverse order.
We then have one matrix embedding for the forward direction $\tX^\fwd \in \R^{n_\mathrm{vocab} \times d \times d}$ and one for the backward direction $\tX^\bwd \in \R^{n_\mathrm{vocab} \times d \times d}$. Then we concatenate forward and backward directions. \Figref{fig:w2m:bidirection} illustrates bidirectional CMOW.

Furthermore, we emit one representation per token position $i$, which allows training with a masked language model objective~\cite{bert}. Thus, we are able to make use of the BERT teacher signal for pretraining. Since we can reuse computations, $\mathcal{O}(n)$ matrix multiplications are sufficient to encode a sequence of length $n$.
For these intermediate representations, we also modify the CBOW component in a way that it yields partial sums for the forward and backward directions.
Formally, we compute the \CMOWCBOWHybrid representation as follows: 

\begin{align*}
  \mH^\bidicmow_i &:= \mX_1^\fwd 
  \cdots\mX_i^\fwd \cat{} \mX_{n}^\bwd 
  \cdots\mX_{i}^\bwd\\
  \vh^\bidicbow_i &:= \sum_{j=1}^i \vx_j^\cbow \cat{} \sum_{j=i}^n \vx_j^\cbow\\
    \vh^\bidihybrid_i &:= \operatorname{flatten}\left(\mH^\bidicmow_i\right) \cat{} \vh^\bidicbow_i
\end{align*}

For fine-tuning on tasks with full sentences as input, \eg natural language inference, we do not need per-token representations. 
In this case, we compute the representation of the full token sequence as follows:
\begin{align*}
  \mH^\bidicmow := & \mX_1^\fwd \cdot\mX_2^\fwd \cdots\mX_n^\fwd \cat{} \\
                & \mX_{n}^\bwd \cdot\mX_{n-1}^\bwd \cdots\mX^\bwd_1\\
  \vh^\cbow :=& \sum_{j=1}^n \vx_j^\cbow\\
    \vh^\bidihybrid :=& \operatorname{flatten}\left(\mH^\bidicmow_i\right) \cat{} \vh^\cbow_i
\end{align*}
Note that the forward and backward directions of the embedding vectors $\vh^\cbow$ conflate to equivalent formulas when we encode entire sequences. Thus, we only need to include a single CBOW representation along with the two CMOW components that yield different results for the forward and backward direction. At inference time, the model is parallelizable along the sequential dimension.

For regularization, we apply a mild dropout ($p=0.1$) on both the embeddings and their aggregated representations during pretraining. Then, we feed them into a linear masked language modeling head, see Figure~\ref{fig:w2m:bidirection}, or an MLP classification head to tackle the downstream tasks. 

We have also experimented with linear, LSTM, and CNN classification heads. 
We chose an MLP because it adds non-linearity to the model without introducing further complexity. The MLP has led to the best average performance across all GLUE tasks. We report the detailed results, also with the other downstream classifiers, in the supplementary material.

\subsection{Cross-Architecture Distillation}

A central question of our research is whether we can distill a large PreLM, \eg BERT, into more efficient, non-transformer architectures such as the proposed bidirectional \CMOWCBOWHybrid model.
This requires a cross-architecture distillation approach, which we describe below.

In general, the idea of knowledge distillation is to compress the knowledge of a large teacher model into a smaller student model~\cite{knowledgedistillation,modelcompression}.
It involves a loss function $\Ls$ that is a combination of two loss terms, \ie $\Ls = \alpha \cdot \Lhard + (1-\alpha) \cdot \Lsoft$ with weighting parameter $\alpha$.
$\Lhard$ denotes the cross-entropy loss with respect to the ground truth and $\Lsoft = \Sigma_i \evt_i \cdot \log(\evs_i)$ is the cross-entropy between the student logits $\vs$ 
and the teacher signal $\vt$. Optionally, the softmax within $\Lsoft$ is flattened by a temperature parameter $T$.
We distinguish \emph{general distillation}, where BERT's teacher signal is only used during pretraining, and \emph{task-specific distillation}, where the BERT teacher signal is used during fine-tuning for the downstream task (see \Secref{sec:cad:problem}). 

Considering our goal to design a cross-architecture distillation, the general distillation approach has the conceptual benefit that the teacher model is not needed for fine-tuning. Thus, the student model is capable of tackling downstream tasks without the supervision of the large teacher. 
This has the benefit that one does not need to carry around the BERT model for adaption to every new downstream task.
Above, we have introduced the ability to emit per-token representations with lightweight (bidirectional) \CMOWCBOWHybrid embedding models.
This enables us to use BERT's teacher signal during pretraining together with a masked language modeling objective.
In other words, this enables us to perform cross-architecture distillation with matrix embeddings.

We consider three variants of cross-architecture distillation in our experiments: 
a)~When using general distillation, depicted in Figure~\ref{fig:w2m:bidirection}, BERT acts as a teacher during pretraining and the model is fine-tuned to downstream tasks on its own.
b)~For task-specific distillation, BERT acts as a teacher during fine-tuning as shown in Figure~\ref{fig:w2m:diffcat}. 
For this case, we have the option of either b1)~starting with pretrained embeddings (from general distillation, \ie the a) variant), or b2)~starting from scratch with randomly initialized embeddings.

\subsection{Two-Sequence Encoding with Matrix Embeddings}
\label{sec:diffcat}
When fine-tuning our matrix embeddings to downstream tasks, we can deviate from BERT's input processing, even if BERT is used as a teacher. This is because the distillation loss is computed per sentence (pair) and not per token. The input processing of BERT encodes two sequences by joining them into one sequence.
For example, in a natural language inferencing task, there is a sentence $A$ that potentially entails a sentence $B$, which is encoded together as a single sequence using a special separator token.
This encoding scheme is less useful to our matrix embeddings without any attention component, since the order-aware matrix multiplications would blend the representation of the two sequences.

To develop an appropriate two-sequence encoding scheme for matrix embeddings, we take inspiration from the pre-transformer era, \eg Mou et al.~\cite{DBLP:conf/acl/MouMLX0YJ16}, and from SentenceBERT~\cite{sentenceBert}.
The key idea is to encode two sentences $A$ and $B$ separately before combining them.
As combination operation, we use the absolute elementwise difference and concatenate it to the representations of $A$ and $B$, which we denote as DiffCat:
\begin{align*}
    \vh^{\text{(DiffCat)}} = \vh^\text{(A)} 
    \,\,||\,\, 
    |\vh^\text{(A)} - \vh^\text{(B)}|
    \,\,||\,\, 
    \vh^\text{(B)} 
\end{align*}

\begin{figure*}[htb]
    \centering
    \includegraphics[width=0.75\linewidth]{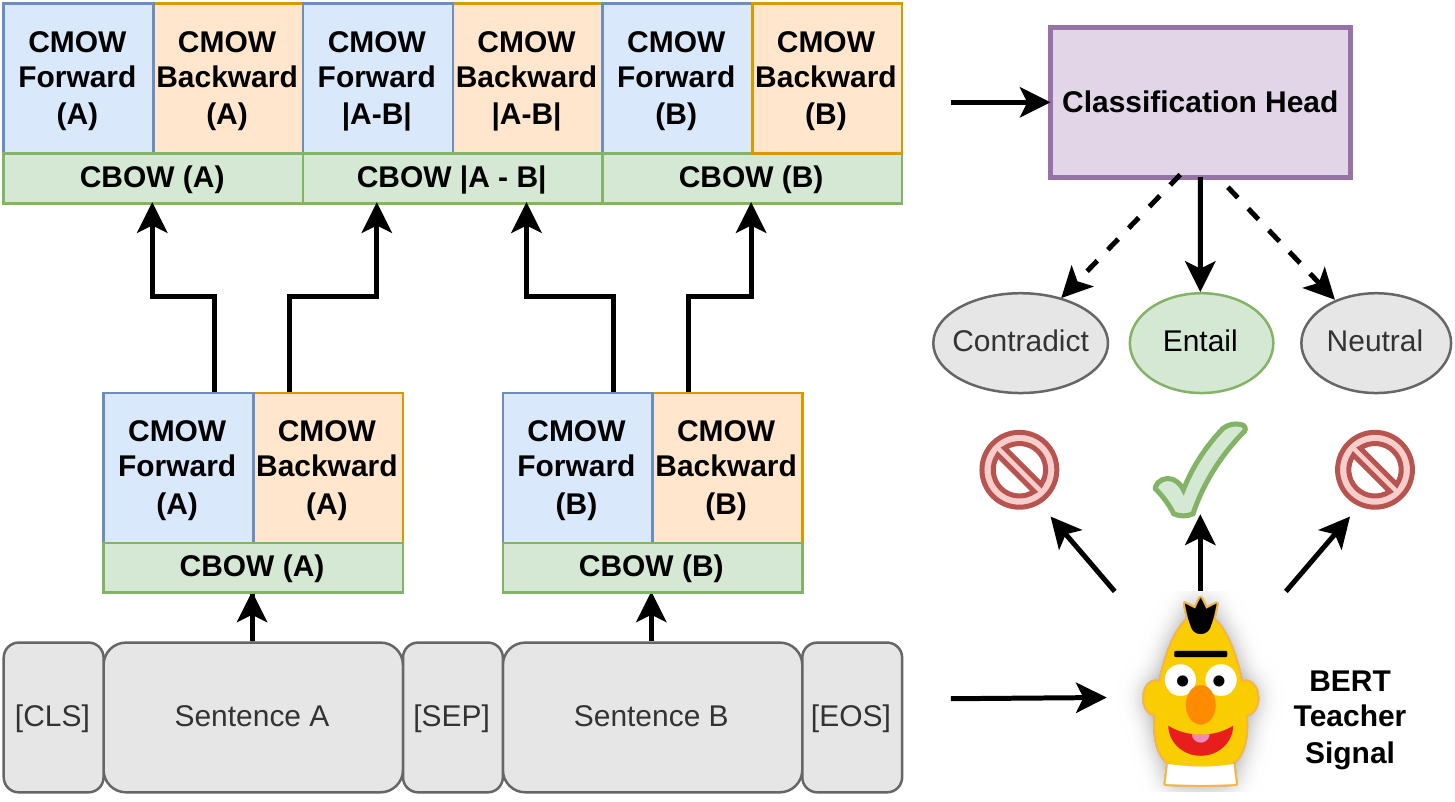}
    \label{fig:w2m:diffcat}
\end{figure*}

We illustrate this separate encoding scheme during task-specific distillation in \Figref{fig:w2m:diffcat}.
The rationale for using a concatenation of both sequence representations along with their difference is that we add a component for the similarity of the two sequence representations, without compromising expressive power.

\section{Experimental Procedure}
\label{sec:apparatus}\label{sec:experimental-procedure}
The experimental procedure is divided into pretraining on unlabeled text and fine-tuning on the downstream tasks.
We provide the details for these two stages and close the section by outlining the downstream tasks and evaluation measures.

\subsection{Pretraining and General Distillation}
In the pretraining stage, as shown in Figure~\ref{fig:w2m:bidirection}, we train our proposed bidirectional \CMOWCBOWHybrid model with a masked language modeling objective (MLM)~\cite{bert} on large amounts of unlabeled text. The MLM objective is to predict left-out words from their context.
We put equal weights on the MLM objective and the teacher signal from BERT ($\alpha=0.5$).
As suggested by Liu~\etal~\cite{roberta2019} and Sanh~\etal~\cite{distilbert}, we do not use the next-sentence prediction objective of BERT, but only the MLM objective. 

As datasets for pretraining, we use a combination of English Wikipedia and Toronto Books~\cite{DBLP:conf/iccv/ZhuKZSUTF15}, as used in the original BERT.
To reduce the environmental footprint of our experiments, we have only pretrained a single bidirectional \CMOWCBOWHybrid model with BERT-base as a teacher on the full unlabeled training data, after pre-experiments on $10\%$ of the training data showed that the selected bidirectional \CMOWCBOWHybrid with distillation exceeded the performance of the baseline.

We use matrix embeddings of size $20 \times 20 = 400$ ($d=20$) for both the CMOW directions (forward and backward) and vector embeddings of size $d_\mathrm{vec} = 400$. 
We use BERT's tokenizer and its vocabulary for both, the teacher and student\extended{~(to ensure both use the same vocabulary)}. 
The BERT tokenizer relies primarily on the WordPiece algorithm~\cite{wordpiece}, which yields a high coverage while maintaining a small vocabulary.

\subsection{Fine-tuning and Task-specific Distillation}
In the fine-tuning stage, as shown in Figure~\ref{fig:w2m:diffcat}, the pretrained model is adapted for each downstream task individually. 
The training objective for fine-tuning is either cross-entropy with the ground truth (in general distillation) or a mixture of the ground truth loss and cross-entropy with respect to the teacher's logits (in task-specific distillation). Again, we put equal weight on ground truth and teacher signal along ($\alpha=0.5$) with unit temperature. 
To facilitate distillation on regression tasks, we follow Raffel~\etal~\cite{T5} and cast STS-B from regression to classification by binning the scores into intervals of $0.2$.
To encode the inputs for two-sequence tasks, we use a sequential encoding similar to BERT and the proposed DiffCat encoding (see Section~\ref{sec:diffcat}).

For task-specific distillation, we employ an uncased BERT-base model\footnote{\url{https://huggingface.co/textattack}} from 
the Huggingface repository that has already been fine-tuned for each task of the GLUE benchmark. 
We have fine-tuned the BERT model ourselves on the tasks
STS-B, where we applied binning, and MNLI, where the pretrained model led to subpar results.
We use the same fine-tuned BERT model as a teacher for all experiments with task-specific distillation.

We seek a fair comparison between the unidirectional \CMOWCBOWHybrid baseline model and our bidirectional model. As such, we allow both models to equally benefit from BERT's teacher signal during fine-tuning.
For our comparisons regarding specific components (two-sentence encoding and bidirectionality), we use random initialization for both models because Mai~\etal~'s pretrained embeddings \cite{cmow} came with a different vocabulary that covered only 53\% of the one of BERT. 
Throughout the other experiments, we initialize our bidirectional \CMOWCBOWHybrid with the pretrained embeddings from general distillation, while we isolate the effect of task-specific distillation in a dedicated experiment.

For hyperparameter optimization, we tune learning rates in the range of $\lbrack 10^{-3}, 10^{-6}\rbrack$. 
In total, we have conducted $306$ training and evaluation runs for hyperparameter optimization of each GLUE task. 
To determine the best model, we use each task's evaluation measure on the development set. We run each model for $20$ epochs with early stopping ($5$ epochs patience). 
We select appropriate batch sizes on the basis of preliminary experiments and training data sizes. 

In the supplementary material, we provide a more detailed discussion of the hyperparameters. We also report the hyperparameter values of the best-performing models.
We have also experimented with data augmentation and using exclusively the teacher signal during task-specific distillation ($\alpha=1$).
In some tasks, we could further increase the results by a small margin, but found no consistent improvement. 
These additional experiments can also be found in the supplementary materials.

\extended{
For learning these embeddings, we have used a combination of the masked language model objective~\cite{bert} and a distillation objective with respect to a pretrained BERT model.}

\extended{The exact bounds and optimization method for every hyperparameter is reported in the supplementary materials.}

\subsection{Downstream Tasks and Measures} \label{task-measures}
We use the GLUE benchmark~\cite{glue} to evaluate our models.
The GLUE benchmark consists of nine tasks for English language comprehension \cite{glue}. These tasks comprise natural language inference (MNLI-m, QNLI, WNLI, RTE), sentence similarity (QQP, STS-B, MRPC), linguistic acceptibility (CoLA), and sentiment analysis (SST-2). All tasks are based on pairs of sentences except CoLA and SST-2, which are single-sentence tasks.
The GLUE benchmark explicitly encourages the use of different fine-tuning strategies for different tasks.
For our evaluation, we use the GLUE development set along with its task-specific measures.
As such, the performance on all four NLI tasks as well as SST-2 is measured with accuracy.
CoLA is evaluated by Matthews correlation coefficient.
Similarity tasks are measured by the average Pearson and Spearman correlation for the STS-B task, and as the average accuracy and $F_1$-score for MRPC and QQP.

\extended{The same metrics have been used for model evaluation in several papers and therefore, allow for comparing our results to those of other works \cite{bert, sun2019patient, cmow, knowledgedistillation, ladabert, tinybert, boundariesbertdistillation}}

\extended{
\todoyellow{moved this former Evaluation paragraph here}
Every model was trained for at most 20 epochs. 
To find the best model out of all epochs, early stopping based on the evaluation measure with a patience of five epochs was applied. 
\extended{For hyperparameter tuning, we report the scores based on one run with fixed seed.}
Ablation studies were performed for each task using a selection of models that performed well during hyperparameter optimization.
For each model, the best learning rate is determined again in each ablation study.

[[[]Fine-tuning and distillations were performed on single V100 GPUs\extended{, provided by the BWUniCluster \footnote{\url{https://www.bwhpc.de/index.php}}.]]]}
}

\section{Results}
\label{sec:results}

We present the results along the design choices introduced in \Secref{sec:methods}, namely bidirection, cross-architecture distillation approaches, and two-sequence encoding scheme.
We compare our best embedding methods with ELMo and BERT distillates from the literature.
Finally, we report the inference times and the number of parameters of the models.

\subsection{DiffCat Encoding vs. Joint Encoding}\label{sub:cad:exp3}
First, we compare the encoding schemes for two-sentence tasks.
On the one hand, we have the BERT-like encoding that encodes the two sentences together, separated by a special token.
On the other hand, we have the proposed DiffCat encoding, which encodes each sentence separately before combining the representations.
For a fair comparison, we use a randomly initialized unidirectional \CMOWCBOWHybrid model under task-specific distillation.

Table~\ref{tab:cad:results:diffcat} shows that DiffCat encoding improves the results consistently with the largest margin on STS-B. 
The most remarkable improvement is the improvement from $18.5$ to $58.6$ in the STS-B sentence similarity task when encoding the input of the sentence pair through DiffCat.
The average improvement across the two-sentence GLUE tasks is 20\%, when the DiffCat encoding is used over a BERT-like joint encoding of the sentence pairs.
In subsequent experiments, we only report scores with DiffCat encoding.
\begin{table*}[ht]
  \centering
  \small
  \caption{DiffCat encoding vs. joint BERT-like encoding. Both variants use randomly initialized unidirectional \CMOWCBOWHybrid embeddings with MLP under task-specific distillation. DiffCat encoding improves the avg. score across two-sentence tasks by 20\%.}
  \label{tab:cad:results:diffcat}
  \begin{tabular}{lcccccccccc}
    \toprule
    \textbf{Two-Sentence Encoding}     &  \textbf{Avg.} &\textbf{MNLI-m} & \textbf{MRPC} & \textbf{QNLI} & \textbf{QQP}  & \textbf{RTE}  & \textbf{STS-B} & \textbf{WNLI} \\
    \midrule
        BERT-like Joint Encoding & 55.8 & 50.0 & 73.0 & 60.4 & 78.6 & 53.8 & 18.5 & \textbf{56.3}\\
        DiffCat Separate Encoding & \textbf{66.8}  &\textbf{62.5} & \textbf{74.3} & \textbf{71.5} & \textbf{86.6} & \textbf{58.1} & \textbf{58.6} & \textbf{56.3}\\
    \bottomrule
  \end{tabular}
\end{table*}

\subsection{Bidirectional vs. Unidirectional \CMOWCBOWHybrid}
To isolate the effect of the bidirectional component, we compare unidirectional \CMOWCBOWHybrid with bidirectional \CMOWCBOWHybrid under equal conditions.
We train both variants from scratch for the downstream tasks, while using a BERT's teacher signal.
Table~\ref{tab:results:bidirection} shows the results of the comparison of unidirectional Hybrid embeddings with the proposed bidirectional Hybrid embeddings.
Bidirection helps on the tasks MNLI, MRPC, QNLI, SST-2, STS-B, and WNLI. On the other tasks, the difference is marginal.
We have an average improvement of $1\%$ of the bidirectional model over the unidirectional model across all tasks of the GLUE benchmark. 

\begin{table*}[ht]
  \centering
 \caption{Bidirectional versus unidirectional \CMOWCBOWHybrid under task-specific distillation with DiffCat encoding.}
  \label{tab:results:bidirection}
  \small
  \begin{tabular}{lcccccccccc}
    \toprule
    \textbf{Model Type}  & \textbf{Score} & \textbf{CoLA} & \textbf{MNLI-m} & \textbf{MRPC} & \textbf{QNLI} & \textbf{QQP} & \textbf{RTE} & \textbf{SST-2} & \textbf{STS-B} & \textbf{WNLI} \\
    \midrule
    Hybrid, rand. init. & 62.5 & \textbf{13.1} & 62.5 & 74.3 & 71.5 & \textbf{86.6} & \textbf{58.1} & 83.1 & 58.6 & 56.3\\
    Bidirectional Hybrid, rand. init. & \textbf{63.2} & 13.0 & \textbf{63.3} & \textbf{75.7} & \textbf{72.6} & 86.1 & 57.4 & \textbf{83.3} & \textbf{59.7} & \textbf{57.7}\\
    \bottomrule
  \end{tabular}
\end{table*}

\subsection{General Distillation vs. Task-Specific Distillation}
Next, we compare general distillation with task-specific distillation.
As shown in Table~\ref{tab:results:distillation}, using general distillation leads to better results for five tasks (MNLI, MRPC, QQP, STS-B, and RTE) compared to task-specific distillation. 
For the other four tasks (CoLA, QNLI, SST-2, and WNLI), task-specific distillation achieves higher scores. 
The average score of general distillation is higher than with task-specific distillation in both pretrained and randomly initialized cases.

\begin{table*}[ht]
  \centering
  \small
  \caption{Comparison of task-specific vs. general distillation using bidirectional \CMOWCBOWHybrid embeddings.} 
  \label{tab:results:distillation}
  \small
  \begin{tabular}{lcccccccccc}
    \toprule
    \textbf{Distillation Type}  & \textbf{Score} & \textbf{CoLA} & \textbf{MNLI-m} & \textbf{MRPC} & \textbf{QNLI} & \textbf{QQP} & \textbf{RTE} & \textbf{SST-2} & \textbf{STS-B} & \textbf{WNLI} \\
    \midrule
    General & \textbf{66.6} & 16.7 & \textbf{66.6} & \textbf{79.7} & 71.7 & \textbf{87.2} & \textbf{61.0} & 82.9 & \textbf{76.9} & 56.3 \\        
    Task-specific, rand. init & 63.2 & 13.0 & 63.3 & 75.7 & \textbf{72.6} & 86.1 & 57.4 & \textbf{83.3} & 59.7 & \textbf{57.7}\\
    Task-specific, pretrained &64.6 & \textbf{23.3} & 61.8 & 75.0 & 72.0 & 86.3 & 59.9 & 82.9 & 62.9 & \textbf{57.7} \\
    \bottomrule
  \end{tabular}
\end{table*}

\subsection{Comparing Our Best Models to the Literature}
Table~\ref{tab:results:best} shows the results of the best bidirectional \CMOWCBOWHybrid variants using any of the three distillation methods considered. 
As described by Wasserblatt~\etal~\cite{boundariesbertdistillation}, a model needs to capture context and linguistic structure to perform well on CoLA.
We doubled the results for CoLA and SST-2 compared to the best previously reported cross-architecture distillation approaches by Wasserblatt~\etal~\cite{boundariesbertdistillation}. 
Our best models scored higher than ELMo~\cite{DBLP:conf/naacl/PetersNIGCLZ18} on the tasks MRPC, QNLI, QQP, RTE, and WNLI. We achieve higher scores than DistilBERT on RTE and WNLI. 

\begin{table*}[ht]
  \centering
  \caption{Comparison of best embedding-based methods (in bold) with methods from the literature on the GLUE validation set. The $\star$-symbol indicates numbers on the official GLUE test set }
  \label{tab:results:best}
  \small
  \begin{tabular}{lcccccccccc}
    \toprule
    \textbf{Method} & \textbf{Score} & \textbf{CoLA} & \textbf{MNLI-m} & \textbf{MRPC} & \textbf{QNLI} & \textbf{QQP} & \textbf{RTE} & \textbf{SST-2} & \textbf{STS-B} & \textbf{WNLI} \\
         \midrule
        BERT-base (our teacher model) & 78.9 & 57.9 & 84.2 & 84.6 & 91.4 & 89.7 & 67.9 & 91.7  & 88.0 & 54.9\\
        ELMo~\cite{DBLP:conf/naacl/PetersNIGCLZ18} &  68.7 & 44.1 & 68.6 & 76.6 & 71.1 & 86.2 & 53.4 & 91.5 & 70.4 & 56.3 \\
        \midrule
        DistilBERT~\cite{distilbert} &  77.0 & 51.3 & 82.2 & 87.5 & 89.2 & 88.5 & 59.9 & 91.3 & 86.9 & 56.3\\
        $\star$ MobileBERT~\cite{sun2020mobilebert} \ & --- & 51.1 & 84.3 & 88.8 & 91.6 & 70.5 & 70.4 & 92.6 & 84.8 & ---\\
        $\star$ TinyBERT (4 layers)~\cite{tinybert} \ & --- & 44.1 & 82.5 & 86.4 & 87.7 & 71.3 & 66.6 & 92.6  & 80.4 & ---\\
        \midrule
        Hybrid~\cite{cmow} \ & --- & --- & --- & --- & --- & --- & --- & 79.6 & 63.4 & --- \\
        Word2rate~\cite{phua2021word2rate} \ & --- & --- & --- & --- & --- & --- & --- & 65.7 & 53.1 & --- \\ 
        \midrule
        CBOW~\cite{boundariesbertdistillation} \ & --- & 10.0 & --- & --- & --- & --- & --- & 79.1 & --- & ---\\
        BiLSTM~\cite{boundariesbertdistillation} \ & --- & 10.0 & --- & --- & --- & --- & --- & 80.7 & --- & --- \\
        Bidi. Hybrid + MLP (ours) & \textbf{68.0} & \textbf{23.3} & \textbf{66.6} & \textbf{80.9} & \textbf{72.6} & \textbf{87.2} & \textbf{61.0} & \textbf{84.0} & \textbf{76.9} & \textbf{59.2} \\
    \bottomrule
  \end{tabular}
\end{table*}

\extended{
\paragraph{Linear Probe vs. MLP vs. CNN}
\todo{keep this?}
The scores among the student classifiers varied more when siamese DiffCat embeddings were used vs. when it was not used. \fix
Considering the average GLUE score, the MLP student classifier performed better by at least $4.8$ points than all other classifiers in combination with all embedding types.
The good performance of the MLP student classifier is also visible when inspecting the best score for every single task: For 7 out of 9 tasks, a model using the MLP student classifier produced the highest score.
\todo[inline]{Extract results for downstrema classifiers (maybe) or ban that to appendix}
When DiffCat is \textit{not} used, the differences\extended{~in performance of the different student classifiers used for the task-specific distillation} are overall smaller.
Looking at the average GLUE score, the difference in performance is never more than $3.2$. All best scores were produced by the MLP classifier, except in combination with CMOW embeddings. There, the CNN student achieves the best score. 

\paragraph{CMOW vs. CBOW vs. \CMOWCBOWHybrid}
\todo{rephrase the paragraph}
For most classifiers, CMOW embeddings produce higher scores than CBOW embeddings, regardless of their initialization (pretrained or random).
Hybrid embeddings can further improve the performance on 5 out of 9 tasks by a small margin.
Word order seems to be not so important for the MRPC and STS-B tasks. This is suggested by the already strong performance of the CBOW model with siamese encoding. 
Our observation is in line with, \eg \citet{DBLP:conf/acl/BaroniBLKC18}, who found that ‘Word Content’ within sentence embeddings is most correlated with downstream task performance, particularly on STS-B.

\paragraph{Pretrained vs. Random\extended{ly Initialized} Embeddings}
\todo{Rephrase this paragraph, merge with above on distillation}
The results show that using pretrained embeddings instead of randomly initialized embeddings for task-specific knowledge distillation does not necessarily lead to higher scores. 
For all types of embeddings, differences between scores resulting from pretrained and randomly initialized embeddings vary depending on the task and the student classifier.
The biggest gain for pretrained embeddings among all model combinations can be found for CoLA. 
}

\begin{table}[htbp]
    \centering
    \caption{Number of parameters and inference time of the models. Inference speed on the trained model computed using an NVIDIA A100-SXM4-40GB card}
    \small
    \begin{tabular}{lcc}
    \toprule
         Model & \# Parameters & Inference speed (sent./sec)  \\
         \midrule
         ELMo & 94M & 1.1k \\
         BERT-base & 109M & 4.6k \\
         DistilBERT-base & 66M & 9.2k\\
         MobileBERT & 25M & 5.5k\\
         TinyBERT (4 layer) & \textbf{14M} & \textbf{30.0k}\\
         Bidi. Hyrid & 37M & \textbf{30.0k}\\
         \bottomrule
    \end{tabular}
    \label{tab:performance}
\end{table}

\subsection{Runtime Performance and Parameter Count}

To compare runtime performance, we generate 1,024 batches with 256 random sequences of length 64 and measure the inference time (no gradient computation) of the models to encode the sequences. As shown in Table~\ref{tab:performance}, both bidirectional \CMOWCBOWHybrid and, notably, TinyBERT are more than 6 times faster than BERT-base and more than 3 times faster than DistilBERT. Bidirectional \CMOWCBOWHybrid uses only half of DistilBERT's parameters.

The inference speed of \CMOWCBOWHybrid could be increased even further because the $\mathcal{O}(n)$ steps to encode a sequence of length $n$ can be parallelized into $\mathcal{O}(\log n)$ \emph{sequential} steps, since matrix multiplication is associative. 

\section{Discussion and Related Work}\label{sec:discussion}
\paragraph{Key Results} We have shown that BERT can be distilled into efficient matrix embedding models during pretraining by emitting intermediate representations. We have also introduced a bidirectional component and a separate two-sequence encoding scheme for CMOW-style models.
We have observed that the general distillation approach, \ie using the BERT teacher only during pretraining, leads to results that are oftentimes even better than those achieved with task-specific distillation. This is an interesting result because all previous works on cross-architecture distillation relied on task-specific distillation. Our proposed model offers an encoding speed at inference time that is three times faster than DistilBERT and more than five times faster than MobileBERT.

\extended{
As future work, different variations of knowledge distillation could be used with our order-aware embeddings as student.
One approach is the gradual soft loss reducing method, where knowledge distillation first uses both, hard and soft losses, but then the weight of the soft loss is reduced further and further during the training process \cite{shin2020knowledge}.
Another approach is online knowledge distillation, which is often used in image classification. 
Hereby, a peer-teaching is applied, where one teacher and many students are trained at the same time with the speciality that the students learn from each other~\cite{lan2018knowledge}.
The combination of distillation and model  compression techniques is also a promising approach \cite{polino2018model, ladabert}. 
Finally, it would be interesting to add an attention module to obtain better scores on CoLA. 
We have tried simple self attention variants but have not pursued this path further because associativity would no longer hold, and thus, efficiency would suffer.
Still it might be an interesting direction of future work.

Finally, as a variation of the teacher assistant approach, the teacher models could be replaced with other models~\cite{tinybert} or trained with early stopping, in order to decrease the size gap between teacher and student.
}

\extended{We summarize the work on continuous text representation models and model compression.
We contrast approaches for model compression based on a general distillation and task-specific distillation~\cite{tinybert}.
We also give a brief overview over pruning and quantization as techniques that can be used to further reduce the model size.}

\extended{
\paragraph{Order-aware Embeddings}
\todo[inline]{Most of the stuff here has been said in the introduction. so it could be omitted here. actually the only new information is almost word2rate. ---
and the info that CBOW is easy to compute should be mentioned in the introduction anyways}
The continuous bag-of-words (CBOW) model~\cite{word2vec} provides dense representations, \ie embeddings of words and their context within a text.
CBOW embeddings are accurate in encoding word content and relatively easy to compute. 
However, they do not capture the word order. 
In contrast, continual multiplication of words (CMOW) embeddings can capture word order, but are less accurate in encoding word content~\cite{cmow}. 
Therefore, the authors propose a \CMOWCBOWHybrid model that combines the advantages of both kinds of embeddings. 
\extended{They found that the hybrid model performs better than CBOW on 8 out of 11 reported downstream tasks.}
Word2rate~\cite{phua2021word2rate} is an  extension of CMOW where the 
matrices are considered as statistical transitions (rate matrices) in a Taylor series, with comparable results. 
}

\extended{
\paragraph{Transformer Models}
In 2017, the Transformer model was introduced by Vaswani et al.~\cite{attention}. 
It requires less computational resources than previous models while outperforming them in English-to-German as well as English-to-French translation tasks~\cite{attention}.
One of the most influential transformer models for natural language processing is BERT~\cite{bert}. 
Its training consists of two steps: First, an unsupervised pretraining step on a large, general data set is performed.
In the next step, the pretrained model is fine-tuned to various types of specific downstream tasks.
}
 
\extended{
\paragraph{Reducing Model Size}
Since BERT contains a large number of parameters and training requires significant computational effort, there have been attempts to reduce it using different techniques, especially pruning~\cite{DBLP:conf/nips/Sanh0R20} and distillation \cite{distilbert, tinybert}.
The latter is a promising method for model compression on which we will focus in the following \cite{modelcompression, knowledgedistillation}: A student learns to mimic the behavior of the teacher. The aim is to achieve comparable accuracy, but with a smaller, and above all faster, model. While hard labels are normally used in supervised learning, for knowledge distillation, soft labels collected from the teacher are used to train the student model. This way, the small residual probabilities that are normally ignored when using hard labels are also taken into account \cite{knowledgedistillation}. 
}

\paragraph{Reflection to Same-Architecture Approaches}
\extended{In a recent extension of CMOW, Word2rate~\cite{phua2021word2rate}, the matrices are considered as statistical transitions (rate matrices) in a Taylor series.
The authors report the scores of Word2rate on SST-2 and STS-B, which are lower than the scores of the original \CMOWCBOWHybrid. }
Recall that in general distillation\label{general_distillation}, a PreLM is distilled into a student model during pretraining. 
DistilBERT~\cite{distilbert} is such a general-purpose language model that has been distilled from BERT. 
Apart from masked language modeling and distillation objectives, the authors also introduced a cosine loss term to align the student's and teacher's hidden states (layer transfer). 
Furthermore, the student is initialized with selected layers of the teacher.
\extended{It was created by performing distillation only during the pretraining phase via the masked language model objective. 
The once-distilled model is then fine-tuned for each downstream task \cite{distilbert}.}
MobileBERT~\cite{sun2020mobilebert} introduced a bottleneck to BERT such that layers can be transferred to student models with smaller dimensions.
\extended{The structure of MobileBERT enriches BERT's structure by using an inverted bottleneck before and a bottleneck after the encoder blocks, making the model deeper and reducing its width.
This makes MobileBERT amenable for layer transfer to the smaller student~\cite{sun2020mobilebert}.}

In task-specific distillation, the teacher signal is used during fine-tuning. 
Sun~\etal~\cite{sun2019patient} use layer-wise distillation objectives and initialize with teacher weights to train BERT students with fewer layers.
\extended{They explore which layers from the teacher supply the most information, \eg using upper layers or learning from every $k$th layer.}
TinyBERT~\cite{tinybert} applies knowledge distillation in both stages, pretraining and fine-tuning\extended{ with BERT-base as teacher}. 
LadaBERT~\cite{ladabert} combines knowledge distillation with pruning and matrix factorization.
Other approaches consider distillation in multi-lingual~\cite{DBLP:conf/emnlp/TsaiRJALA19} or multi-task settings~\cite{yang2019multitaskKD}. 

Bidirectional \CMOWCBOWHybrid yields high throughput rates comparable to a 4-layer TinyBERT~\cite{tinybert}. 
In the original TinyBERT~\cite{tinybert} work, the authors report a speed-up of 2x with 6 layers compared to BERT-base, and 9.4x with 4 layers.
However, TinyBERT requires to have the teacher model available for fine-tuning. 
We have shown that \CMOWCBOWHybrid is better even when using only general distillation compared to using task-specific distillation.
TinyBERT further augments the training data\extended{~\cite{tinybert}}, which we have also considered, but we found no consistent improvement.

Turc~\etal~\cite{turc2019wellread} analyze the interaction between pre-training and fine-tuning with BERT models and find that pretrained distillation works well, which agrees with our findings on the importance of pretraining with CMOW-style models.
\extended{For data augmentation, a single sentence from the corpus is used to generate $n$ augmented sentences.
Single-piece words are masked and predicted again using some pretrained BERT model.
The masked word is replaced by a randomly selected candidate whose probability is greater $0.4$. 
For all other words, new candidates are generated using n-closest GloVe word embeddings based on cosine similarity. 
This procedure is repeated for every word in a sentence to get the new augmented sentence. Data augmentation is applied 20 times per sentence \cite{tinybert}.}

\paragraph{Reflection w.r.t. Cross-Architecture Approaches}
The distillations of BERT described above assume that the teacher and student share the same architecture.
\extended{This allows to use techniques such as layer transfer and loss terms to align hidden states.}
However, the student model does not need to have the same architecture as the teacher, which is what we call cross-architecture distillation.
For example, Wasserblatt~\etal~\cite{boundariesbertdistillation} use a simple feed forward network with CBOW embeddings and a bidirectional LSTM model as students. 
Both models perform well in several downstream tasks.
Tang~et al.~\cite{tang2019distilling} explore the distillation of BERT into a single-layer BiLSTM without using additional training data or modifications to the teacher architecture. 
Their distillation-based approach yields improvements compared to a plain BiLSTM without teacher signal: about 4 points on all reported tasks (QQP, MNLI, and SST-2).
This has motivated us to investigate whether even more efficient models can be used as students of a BERT teacher.

\paragraph{Reflection w.r.t. Pruning and Quantization}
Other techniques for reducing the size of a model are pruning and quantization.
Pruning approaches such as in Sanh~\etal~\cite{DBLP:conf/nips/Sanh0R20} reduce the number of parameters. 
Still, the resulting smaller models use the same architecture as their larger counterparts and, thus, pruning does not necessarily improve inference speed.\extended{ until dedicated hardware for sparse networks becomes available~\cite{DBLP:conf/nips/Sanh0R20}.}
Quantization is a common post-processing step to reduce model size by decreasing the floating point precision of the weights~\cite{DBLP:conf/icml/GuptaAGN15,wu2020integer}. 
Pruning and quantization can be applied in conjunction with knowledge distillation~\cite{DBLP:conf/nips/Sanh0R20,sun2020mobilebert}. 
Aside from techniques for reducing the model size, there is also a tremendous effort to improve the efficiency of the transformer architecture in the first place \cite{efficient-transformers}.

\extended{
The literature focuses on reducing the size of PreLMs via distillation, pruning, and quantization.
Specialized techniques depend on teacher and student sharing the same architecture, and thus are not applicable for cross-architecture distillation.
So far, only few recent works consider distilling PreLMs into other architectures like LSTMs and CBOW models. 
In this work, we show that cross-architecture distillation with order-aware embedding models as students outperform previous cross-architecture distillation approaches and achieve scores competitive to ELMo, while using less computational resources.
\todo{check publications since ICLR}
Finally, all previous cross-architecture distillation approaches are task-specific, \eg Wasserblatt et al~\cite{boundariesbertdistillation}, while we could show that general distillation can lead to even higher scores than task-specific distillation.
}

\paragraph{Threats to Validity}
For a fair comparison, we have ensured that our baselines and the proposed extensions have equal conditions. When testing the effect of bidirectional component and the separate-encoding scheme, we start with random initialization with for the baseline model and for the extended models. Moreover, we have put the same effort into hyperparameter tuning for the baselines and the proposed extensions, as well as when comparing general and task-specific distillation with the proposed model.

Currently, cross-architecture distillation approaches still fall behind other BERT distillates, such as MobileBERT and TinyBERT, in many of the downstream tasks. In particular, detecting linguistic acceptability remains a challenge for non-transformer methods, even though we improve upon previous cross-architecture distillation approaches. So far, we have not analyzed the trade-off between embedding size and downstream performance. We rely on general arguments for the benefits of increased dimensionality~\cite{DBLP:conf/iclr/WietingK19}.

\paragraph{Future Work} \extended{Our work demonstrates how large PreLMs can be distilled into order-aware embedding models.}
One could further improve the efficiency by applying pruning~\cite{DBLP:conf/nips/Sanh0R20} and/or quantization~\cite{wu2020integer} techniques on the learned matrices to allow sparse matrix multiplication during encoding.
Future work could explore what components would be necessary to improve scores on particularly challenging downstream tasks such as detecting linguistic acceptability.
This might include the introduction of a small attention module like that of gMLP~\cite{liu2021pay} to \CMOWCBOWHybrid.
\extended{Since we have shown how to emit (and train) per-token representations with CMOW, an interesting direction of future work would be to explore whether CMOW representations are suited as order-aware embeddings in transformer models, which might alleviate the need for a dedicated position embedding.
Similarly, per-token representations enable tackling question answering tasks with CMOW-style models, in which a span of the context should be predicted.}
Finally, matrix embeddings could also be used in other domains where sequences of discrete elements need to be encoded, \eg events, or gene sequences.

\extended{Further, we note how order-aware embeddings can be implemented and scaled efficiently, as scaling issues that apply to transformer-based architectures do not apply to our architecture.}

\section{Conclusion} 

This research contributes to the development of simpler and more efficient models for natural language processing in general. 
We have introduced three extensions to the CMOW/CBOW-Hybrid model: a bidirectional component, a separate two-sequence encoding scheme, and the ability to emit per-token representations. These per-token representations allow us to distill BERT into CMOW/CBOW-Hybrid already during pretraining with a masked language modeling objective.
Our results show that a separate encoding scheme improves the performance of CMOW/CBOW-Hybrid on two-sentence GLUE tasks by 20\%, while bidirection improves the performance by 1\% compared to the unidirectional model.
Furthermore, we have shown that general distillation seems to be sufficient, and task-specific distillation is not necessary for most GLUE tasks.  
In comparison to more expensive language models reported in the literature, our embedding-based approach achieves scores that match or exceed the scores of ELMo and are competitive to DistilBERT on QQP and RTE with only half of its parameters and thrice its encoding speed.
While linguistic acceptability remains a challenge for non-transformer models, our approach yields notably higher scores than previous cross-architecture distillation approaches.

\section*{Ethics Statement}
The drawback of PreLMs, as discussed in the introduction, is that they are growing larger and larger, which leads to higher environmental and economic costs~\cite{DBLP:conf/acl/StrubellGM19} and shifts the development and training of PreLMs into the hands of a few global players~\cite{foundation-models}.
Large language models have also been criticized for having biases because they are trained on a large amount of weakly curated training data~\cite{DBLP:conf/fat/BenderGMS21}. 
Moreover, in vision, the reduction of model sizes has also been reported to amplify bias~\cite{hooker2020characterising}.
This is particularly a problem when the model is not explainable.
In our proposed method, we distill large-scale PreLMs into models that do not have a nonlinearity (except for the final classification head). 
Since linear models are easier to explain, we hope to contribute to more explainable language models with our work, even though explainability is not the focus of the present work. In fact, models conceptually similar to the ones presented here have been used to explain black-box representations~\cite{DBLP:conf/blackboxnlp/SoulosMLS20}.

\section*{Reproducibility Statement}

We provide the code and pretrained models: 
\href{https://github.com/lgalke/cross-architecture-distillation}{https://github.com/lgalke/cross-architecture-distillation}
Furthermore, supplementary material is available in the appendix of this version, including extended results.

\extended{To ensure that our experiments are reproducible, we have tracked all hyperparameters throughout experimentation. We have described all steps for data processing, \ie tokenization (we use the BERT tokenizer) and binning (for STS-B) in the main part. The datasets from the GLUE benchmark are
well-known to the community, because of which we have omitted a detailed description, except for the metrics that we have reported.
}

\arxivonly{\subsubsection*{Acknowledgement} 
The authors acknowledge support by the state of Baden-Württemberg through bwHPC.
\creditmasterproject{WS 2020/21}\mysupervisorrole}

\bibliographystyle{IEEEtran}
\bibliography{bibliography}

\clearpage
\appendix

\input{appendix}

\end{document}

%% file: appendix.tex
\subsection{Overview of Architectural Choices}\label{app:overview}
\begin{figure*}[!th]
    \centering
    \includegraphics[width=0.9\textwidth]{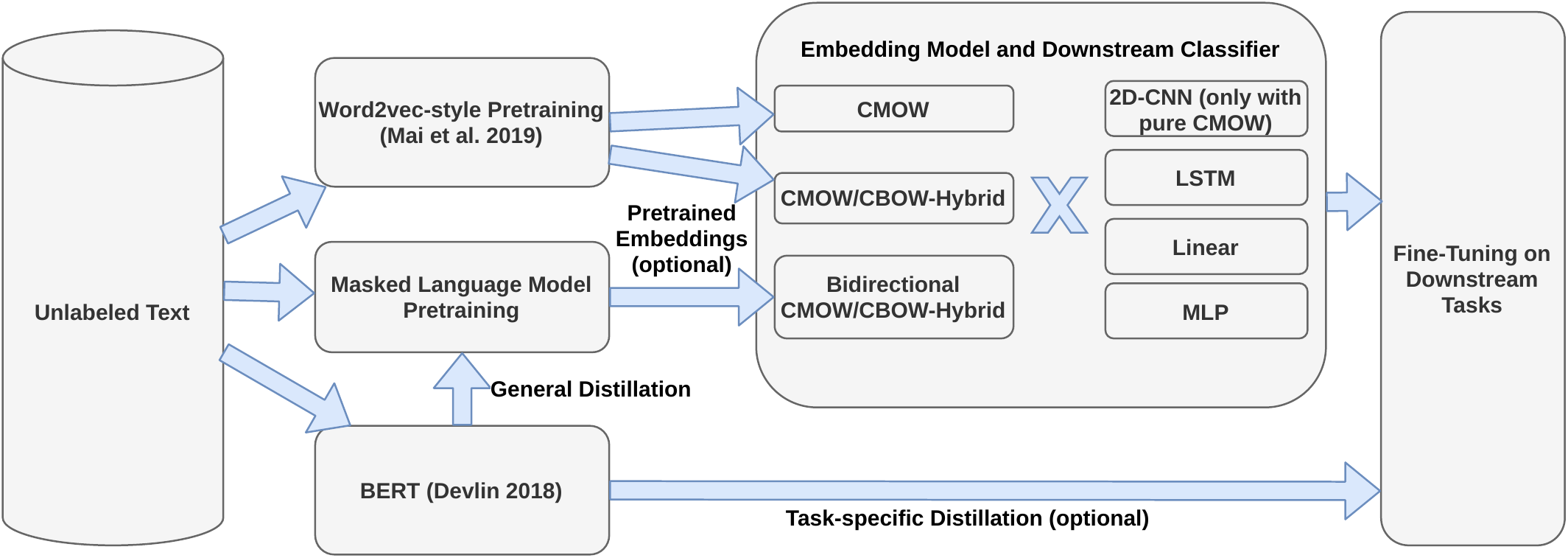}
    \caption{All considered for embeddings and downstream classifiers, pretraining and fine-tuning.}
    \label{fig:overview}
\end{figure*}

\Figref{fig:overview} provides an overview of the architectural choices explored in this paper. We use pretrained BERT~\cite{bert} as well as the embeddings from Mai~\etal~\cite{cmow} as teacher for general distillation. Additionally, we pretrain a model CMOW/CBOW-Hybrid with our extension of masked language model training and bidirection on the same English Wikipedia + Toronto Books dataset. These pretrained embeddings may serve as initialization for the downstream classification models. We also evaluate downstream classification models that have been initialized randomly. For the downstream classification models, we consider three types of embeddings along with four types of classifiers.

For training on the downstream task, we use once again BERT as a teacher for task-specific distillation, while our experiments on general distillation only benefit from the initalization of the MLM-pretrained CMOW/CBOW-Hybrid model.
Throughout the main part of the paper, we have reported scores with an MLP downstream classifier, which achieved the highest average scores.
We have further experimented with a linear downstream classifier, LSTM, and 2D-CNN, which we briefly describe below.

\paragraph{LSTM}
We have further experimented with pooling the sequence of embeddings with an LSTM.
In the past, BiLSTM models have been successfully used in sentiment analysis tasks~\cite{sent_analysis_bilstm,single_bilstm}.
In an LSTM network, the information at hand is propagated in the forward direction. 
Thus, each state $t$ depends on its predecessor $t-1$.
BiLSTM are LSTM networks, in which the inputs are processed twice: once in the forward direction and once in the backward direction, generating a set of two outputs.
In order to generate the output vectors, the output of a single BiLSTM block is fed into an MLP, consisting of two consecutive linear layers with ReLU activation functions.
Note that the BiLSTM operates on a sequence of token embeddings, instead of operating on pooled sentence embeddings like the other student models.
We apply a dropout of $0.5$ after the first linear layer.

\paragraph{CNN}
%
We also explore a 2D-CNN classifier that induces a bias for learning two-dimensional structures within the (aggregated) embedding matrices. 
The CNN consists of one transposed convolution, which increases the matrix dimensions by a factor of four.
Following that, we employ a block of three convolutional layers, the first one having a single filter (or two, for hybrid variants) and a kernel size of four, with the remaining two layers having 3 (4) kernels with stride 2.
To avoid distorting the input embeddings, no padding is applied.
ReLU is used for all activation functions.
We apply BatchNorm for regularization before the last convolutional layer's output is flattened and passed into a linear layer, which produces the predictions.
We add a dropout of $0.4$ before the last linear layer.

\subsection{Discussion of Hyperparameters and Loss Functions for Distillation}\label{app:hyperparameters}
\begin{table*}[]
\small
    \centering
    \caption{Hyperparameter search space and optimization method}
    \begin{tabular}{lcr}
    \toprule
         \textbf{Hyperparameter} & \textbf{Range} & \textbf{Opt. method}\\
         \midrule
         \multicolumn{3}{c}{\textit{--- General Distillation ---}}\\
         Learning rate & $\{10^{-3}, 5\cdot 10^{-4}, 10^{-4}, 5\cdot 10^{-5}, 10^{-5}\}$ & grid search\\
         Warmup steps & $\{0, 500\}$ & grid search\\
         Embedding dropout & $\{0, 0.1\}$ & grid search\\
         Hidden unit dropout & $\{0.2\}$ & fixed\\
         Batch size & \{1,8,32,64,128,256\} & manual\\
         \midrule
         \multicolumn{3}{c}{\textit{--- Task-specific Distillation ---}}\\
        Learning rate & $\{10^{-3}, 5\cdot 10^{-4}, 10^{-4}, 5\cdot 10^{-5}, 10^{-5}, 5\cdot 10^{-6}\}$ & grid search\\
        Embedding type & Hybrid, CMOW, CBOW & grid search\\
        Embedding initialization & random, pretrained & grid search\\
        DiffCat & true, false & grid search\\
        Bidirectional & true, false & grid search \\
        Classifier & Linear Probe, MLP, CNN, BiLSTM & grid search\\
         \bottomrule
    \end{tabular}
    \label{tab:hparams}
\end{table*}
We list hyperparameter search spaces along with their optimization methods in Table~\ref{tab:hparams}.
For the experiments on data augmentation and using only soft loss, we keep the configurations of the best models (See Table~\ref{tab:best_hparams} and tune the learning rate, again.
We optimize over all six initial learning rates, namely $\{ 10^{-3}$, $5\cdot 10^{-4}$, $10^{-4}$, $5\cdot 10^{-5}$, and $10^{-5}\}$. All initial learning rates decay linearly over the course of training.
Note, we also experimented with using warmup steps versus no warmup for the learning rate schedule.
As the warmup did not improve the results, we did not use it.

For the softmax temperature, we find that $T=1$ is often used \cite{knowledgedistillation, ladabert, tinybert, mishra2017apprentice, polino2018model}. 
Since a higher temperature also flattens the curve over all predictions, it could add too much noise and it is therefore better to use a smaller temperature \cite{chen2017learning}.
Setting the weight $\alpha=1$ corresponds to only using hard loss and $\alpha=0$  to only using soft loss. 
Since we do not want to discard any information stemming from the hard loss, we do not follow the approach of Wasserblat et al. who only use the soft loss \cite{boundariesbertdistillation} but instead, we employ a vanilla knowledge distillation approach following Hinton et al.~\cite{knowledgedistillation}.

Hinton~\etal~\cite{knowledgedistillation} state, that using cross-entropy loss on the softmax-temperature with a
large temperature, for example $T=20$, corresponds to only using the Mean Square Error (MSE) loss on the raw student and teacher logits. 
Therefore it is also common to use this loss for the soft distillation loss~\cite{tang2019distilling, boundariesbertdistillation, mukherjee2020distilling}. 
While Tang et al.~\cite{tang2019distilling} used the weighted hard cross-entropy loss in the overall loss calculation, Wasserblat et al. and Mukherjee et al. 
only used the soft loss \cite{tang2019distilling, boundariesbertdistillation, mukherjee2020distilling}. 
A disadvantage of MSE loss is that every error has a huge effect on the overall loss, since it is squared. 
Another point is, that Hinton et al. found it beneficial to use small temperature values if the teacher is way bigger than the student \cite{knowledgedistillation}. 
Since using MSE loss corresponds to using big $T$ values, this loss does not apply to our use case of using small students for lower bound knowledge distillation, 
but with cross-entropy loss, we still have the possibility to achieve the behavior of the MSE loss by setting the value of $T$ to a big value.

\subsection{Extended Results}\label{app:extendedresults}
In the following, we provide extended results for task-specific distillation including the different downstream
classifiers, unidirectional CMOW/CBOW-Hybrid, and joint two-sequence encoding.
The best performing model per task are marked in bold.
We abbreviate CMOW/CBOW-Hybrid as 'Hybrid'.

For the unidirectional baseline model \CMOWCBOWHybrid, we initialize with pretrained embeddings provided by Mai~\etal~\cite{cmow}\footnote{Downloaded from Zenodo: \url{https://zenodo.org/record/3933322\#.YKJ\_uxKxXJU}}, which cover 54\% of BERT's vocabulary. 
As initialization for the newly developed bidirectional \CMOWCBOWHybrid models, we use our own pretrained model obtained by general distillation with BERT.

%
Table~\ref{tab:best_hparams} summarizes the best performing models along with their hyperparameter configurations for each task. Note that we have chosen to use an MLP downstream classifier for the results reported in the main part of this work. Using an MLP downstream classifier has led to the highest average scores across all GLUE tasks.

In Table~\ref{tab:results}, we report an extended version of the comparison with the literature.
Here, we also include our BERT-base teacher model, aswell as TinyBERT~\cite{tinybert} and Tang~\etal~\cite{tang2019distilling}'s distilled BiLSTM. Note that TinyBERT and BiLSTM are not fully comparable, because those numbers are reported on the official GLUE test set, while we have used the validation set for our experiments.

In Table~\ref{tab:results:glue}, we report the results for all downstream classifiers without the DiffCat aggregation but with a sequential BERT-like two-sequence encoding. It is interesting to see that the CMOW-only variant with 2D-CNN classifier leads to the best scores on sentiment analysis task SST-2.
Note that all CMOW variants reported in this table are unidirectional and use task-specific distillation.

In Table~\ref{tab:results:sia}, we report the results for all downstream classifiers with DiffCat two-sequence encoding. Here we observe, that pretrained CBOW with an MLP classifier leads to the best results on sentence similarity (STS-B). Again, all CMOW variants reported in this table are unidirectional and use task-specific distillation.

In Table~\ref{tab:results:bidi}, we report the results for bidirectional models with DiffCat two-sequence encoding.

From all tables combined, we see that Bidirectional CMOW/CBOW-Hybrid model leads to the highest scores on average, even though, on individual tasks, some other variations of the approach lead to higher scores. Thus, we regard bidrectional CMOW/CBOW-Hybrid as our primary model, whose scores we have reported in the main paper, while isolating the effect of the individual components (bidirection, DiffCat encoding, distillation strategies).

We list the number of parameters in Tables~\ref{tab:parameters} and \ref{tab:parameters:sia}. While the absolute numbers might seem high, it is important to note that we have also counted the parameters of the embeddings. As we show in the tables, the number of parameters in the classification models is much lower.

We have performed further experiments with the best performing model for each task: data augmentation and using only soft loss.

\paragraph{Using Only Soft Loss}
We study the influence of the alpha value used in the loss function, based on the best results obtained with the initial $\alpha=0.5$. 
The goal is to investigate whether using only soft loss, \ie setting $\alpha=0.0$ leads to different results. 
As Table \ref{tab:results} shows, using only soft loss improves only the MRPC task by a small margin. 

\paragraph{Data Augmentation}
We conduct a further study with data augmentation as in TinyBERT~\cite{tinybert}. We employ their technique of replacing words by similar word embeddings and nearest predictions from BERT to augment the GLUE training datasets.
The results are shown in Table \ref{tab:results}. 
We find that the effect of data augmentation is small. 
An improvement was only observed on SST-2 ($+1.2$ points) and STS-B ($+3.6$).

\begin{table*}[htbp]
\small
    \centering
    \caption{Hyperparameter configurations for best-performing models by GLUE task}
    \begin{tabular}{lccccccc}
    \toprule
         \textbf{Task} & \textbf{Score} & \textbf{Classifier} & \textbf{Emb. type} & \textbf{Emb. initialization} & \textbf{DiffCat} & \textbf{Bidirectional} & \textbf{Learning rate}\\
         \midrule
        CoLA & {23.3} & MLP & CMOW/CBOW-Hybrid & pretrained & true & true & 1.0E-4\\
        MNLI-m & {63.3} & MLP & CMOW/CBOW-Hybrid & not pretrained & true & true & 1.0E-4\\
        MRPC & {78.2} & MLP & CBOW & pretrained & true & false & 1.0E-3\\
        QNLI & {72.6} & MLP & CMOW/CBOW-Hybrid & not pretrained & true & true & 5.0E-5\\
        QQP & {86.6} & MLP & CMOW/CBOW-Hybrid & not pretrained & true & false & 1.0E-4\\
        RTE & {59.9} & MLP & CMOW/CBOW-Hybrid & pretrained & true & true & 5.0E-4\\
        SST-2 & {86.8} & CNN & CMOW & not pretrained & false & false & 5.0E-4\\
        STS-B & {66.0} & MLP & CBOW & pretrained & true & false & 1.0E-4\\
        WNLI & {69.0} & CNN & CMOW & pretrained & false & false & 1.0E-5\\
         \bottomrule
    \end{tabular}
    \label{tab:best_hparams}
\end{table*}

\begin{table*}[htbp]
    \centering
    \small
    \caption{Scores on the GLUE development set. Our best performing general distillation and task-specific distillation models are highlighted in bold font per task. References indicate sources of scores. The $\star$-symbol indicates numbers on the official GLUE test set. CMOW/CBOW-Hybrid is abbreviated as 'Hybrid'.}
    \label{tab:results}
    \begin{tabular}{lcccccccccc}
        \toprule
         & Score & CoLA & MNLI-m & MRPC & QNLI & QQP & RTE & SST-2 & STS-B & WNLI \\
        \midrule
        \multicolumn{11}{c}{\textit{--- large-scale pre-trained language models ---}}\\
        ELMo~\cite{distilbert} &  68.7 & 44.1 & 68.6 & 76.6 & 71.1 & 86.2 & 53.4 & 91.5 & 70.4 & 56.3 \\
        BERT-base~\cite{distilbert} &  79.5 & 56.3 & 86.7 & 88.6 & 91.7 & 89.6 & 69.3 & 92.7 & 89.0 & 53.5 \\
        BERT-base (our teacher model) & 78.9 & 57.9 & 84.2 & 84.6 & 91.4 & 89.7 & 67.9 & 91.7  & 88.0 & 54.9\\
        Word2rate~\cite{phua2021word2rate} \ & --- & --- & --- & --- & --- & --- & --- & 65.7 & 53.1 & --- \\
        \midrule
        \multicolumn{11}{c}{\textit{--- general distillation baselines ---}}\\
        DistilBERT~\cite{distilbert} &  77.0 & 51.3 & 82.2 & 87.5 & 89.2 & 88.5 & 59.9 & 91.3 & 86.9 & 56.3\\
        $\star$ MobileBERT~\cite{sun2020mobilebert} \extended{w/o OPT}   \ & --- & 51.1 & 84.3 & 88.8 & 91.6 & 70.5 & 70.4 & 92.6 & 84.8 & ---\\
           \midrule
        \multicolumn{11}{c}{\textit{--- task-specific distillation baselines ---}}\\
        $\star$ TinyBERT (4 layers)~\cite{tinybert} \ & --- & 44.1 & 82.5 & 86.4 & 87.7 & 71.3 & 66.6 & 92.6  & 80.4 & ---\\
        CBOW-FFN \cite{boundariesbertdistillation} \ & --- & 10.0 & --- & --- & --- & --- & --- & 79.1 & --- & ---\\
        BiLSTM \cite{boundariesbertdistillation} \ & --- & 10.0 & --- & --- & --- & --- & --- & 80.7 & --- & ---\\
         \midrule
        \multicolumn{11}{c}{\textit{--- general distillation (ours) ---}}\\
       Bidi. Hybrid + Linear  & 65.1 & 15.0 & 63.6 & \textbf{80.9} & 70.7 & 84.3 & 56.7 & \textbf{84.0} & 71.1 & \textbf{59.2} \\        
       Bidi. Hybrid + MLP  & 66.6 & \textbf{16.7} & \textbf{66.6} & 79.7 & \textbf{71.7} & \textbf{87.2} & \textbf{61.0} & 82.9 & \textbf{76.9} & 56.3 \\        
       \midrule
        \multicolumn{11}{c}{\textit{--- task-specific distillation (ours) ---}}\\
        CMOW + CNN (rand. init.) & 54.6 & 13.4 & 45.6 & 72.3 & 61.2 & 82.6 & 56.3 & \textbf{86.8} & 15.0 & 57.8\\
        CMOW + CNN (pretrained) & 56.2 & 18.3 & 50.1 & 71.8 & 60.5 & 80.6 & 57.0 & 85.0 & 13.2 & \textbf{69.0}\\
        CBOW + MLP (pretrained) & 63.8 & 14.0 & 61.7 & \textbf{78.2} & 70.8 & 86.2 & 57.4 & 83.8 & \textbf{66.0} & 56.3 \\
        Hybrid + MLP (rand. init.) & 62.5 & 13.1 & 62.5 & 74.3 & 71.5 & \textbf{86.6} & 58.1 & 83.1 & 58.6 & 56.3\\
        Bidi. Hybrid + MLP (rand. init.) & 63.2 & 13.0 & \textbf{63.3} & 75.7 & \textbf{72.6} & 86.1 & 57.4 & 83.3 & 59.7 & 57.7\\
        Bidi. Hybrid + MLP (pretrained) &64.6 & \textbf{23.3} & 61.8 & 75.0 & 72.0 & 86.3 & \textbf{59.9} & 82.9 & 62.9 & 57.7 \\
        \midrule
        \multicolumn{11}{c}{\textit{--- further experiments on best-performing task-specific distillation models ---}}\\
        Only soft loss ($\alpha=0$) & 64.0 & 19.9 & 62.3 & 78.7 & 72.4 & 68.5 & 56.3 & 86.6 & 62.4 & 69.0\\
        Data augmentation & 63.5 & 21.2 & 47.3 & 76.2 & 72.1 & 86.6 & 52.7 & 88.0 & 69.6 & 57.7\\
        \bottomrule
    \end{tabular}
\end{table*}

\begin{table*}[p]
    \centering
    \small
    \caption{Scores on the GLUE development set without DiffCat encoding}
    \label{tab:results:glue}
    \begin{tabular}{lcccccccccc}
        \toprule
        Task-Specific Distillation & Score & CoLA & MNLI-m & MRPC & QNLI & QQP & RTE & SST-2 & STS-B & WNLI \\
        \midrule
        \multicolumn{11}{c}{\textit{--- task-specific finetuning (ours) ---}}\\
        Teacher BERT-base & 78.9 & 57.9 & 84.2 & 84.6 & 91.4 & 89.7 & 67.9 & 91.7  & 88.0 & 54.9\\
        \midrule
        \multicolumn{11}{c}{\textit{--- task-specific distillation (ours) CBOW not pretrained ---}}\\
        Linear probe & 52.8 & 12.2 & 43.0 & 72.3 & 60.1 & 74.8 & 55.6 & 82.8 & 17.7 & 56.3 \\
        MLP & 53.2 & 13.0 & 46.3 & 71.3 & 59.7 & 76.9 & 54.5 & 82.9 & 17.5 & 56.3\\
        CNN & 52.8 & 11.7 & 43.0 & 72.1 & 60.1 & 77.5 & 54.5 & 82.7 & 17.2 & 56.3\\
        BiLSTM & 52.1 & 10.9 & 44.9 & 70.8 & 59.8 & 78.1 & 54.5 & 81.3 & 12.3 & 56.3\\
        \midrule
        \multicolumn{11}{c}{\textit{--- task-specific distillation (ours) CBOW pretrained ---}}\\
        Linear probe & 52.4 & 11.0 & 43.2 & 72.1 & 58.8 & 74.8 & 54.9 & 82.5 & 14.0 & 60.6\\
        MLP & 54.0 & 14.3 & 46.3 & 71.3 & 60.1 & 76.9 & 58.5 & 83.1 & 14.8 & 60.6 \\
        CNN & 53.0 & 12.0 & 43.5 & 71.6 & 59.2 & 77.5 & 55.2 & 82.6 & 18.8 & 56.3\\
        BiLSTM & 50.8 & 0 & 44.9 & 71.3 & 59.4 & 78.0 & 54.0 & 81.0 & 12.0 & 56.3\\
        \midrule
        \multicolumn{11}{c}{\textit{--- task-specific distillation (ours) CMOW not pretrained ---}}\\
        Linear probe & 53.7 & 13.8 & 45.3 & 72.1 & 62.5 & 
        80.9 & 53.4 & 84.1  & 15.2 & 56.3\\
        MLP & 54.8 & 15.1 & 45.6 & 72.8 & 60.6 & 82.6 & 55.6 & 84.3 & 20.0 & 56.3\\
        CNN & 54.6 & 13.4 & 45.6 & 72.3 & 61.2 & 82.6 & 56.3 & \textbf{86.8} & 15.0 & 57.8\\
        BiLSTM & 53.2 & 16.7 & 44.9 & 72.1 & 64.8 & 80.6 & 54.2 & 82.9 & 7.9 & 54.9\\
        \midrule
        \multicolumn{11}{c}{\textit{--- task-specific distillation (ours) CMOW pretrained ---}}\\
        Linear probe & 54.3 & 20.8 & 48.6 & 71.3 & 60.3 & 78.4 & 54.9 & 84.5 & 13.8 & 56.3\\
        MLP & 55.4 & 18.9 & 50.4 & 72.3 & 61.3 & 79.3 & 55.2 & 83.0 & 17.9 & 60.6\\
        CNN & 56.2 & 18.3 & 50.1 & 71.8 & 60.5 & 80.6 & 57.0 & 85.0 & 13.2 & \textbf{69.0}\\
        BiLSTM & 51.4 & 0 & 44.2 & 68.4 & 59.8 & 81.1 & 55.2 & 82.3 & 15.0 & 56.3 \\
        \midrule
        \multicolumn{11}{c}{\textit{--- task-specific distillation (ours) CMOW/CBOW-Hybrid not pretrained ---}}\\
        Linear probe & 54.4 & 17.0 & 47.0 & 72.6 & 61.1 & 81.4 & 53.4 & 84.5 & 15.1 & 57.8 \\
        MLP & 54.4 & 13.8 & 50.0 & 73.0 & 60.4 & 78.6 & 53.8 & 84.9  & 18.5 & 56.3\\
        CNN & 53.6 & 12.0 & 42.1 & 72.6 & 60.9 & 79.6 & 52.7 & 85.7 & 16.3 & 60.6\\
        BiLSTM & 52.4 & 0 & 43.2 & 72.1 & 61.2 & 80.0 & 57.4 & 83.0 & 18.1 & 56.3\\
        \midrule
        \multicolumn{11}{c}{\textit{--- task-specific distillation (ours) CMOW/CBOW-Hybrid pretrained ---}}\\
        Linear probe & 53.9 & 19.1 & 41.0 & 71.8 & 57.6 & 78.7 & 57.8 & 83.7 & 16.2 & 59.2\\
        MLP & 55.3 & 22.1 & 47.4 & 71.6 & 60.0 & 79.5 & 57.8 & 84.1 & 18.1 & 56.3\\
        CNN & 54.0 & 20.7 & 44.5 & 71.8 & 59.9 & 79.7 & 54.9 & 85.9 & 9.9 & 59.1\\
        BiLSTM & 53.7 & 17.0 & 40.6 & 71.8 & 61.3 & 80.3 & 57.4 & 82.5 & 14.0 & 59.2\\
        \bottomrule
    \end{tabular}
    
\end{table*}

\begin{table*}[p]
    \centering
    \small
    \caption{Scores on the GLUE development set with DiffCat two-sequence encoding}
    \label{tab:results:sia}
    \begin{tabular}{lcccccccccc}
        \toprule
        Task-Specific Distillation & Score & CoLA & MNLI-m & MRPC & QNLI & QQP & RTE & SST-2 & STS-B & WNLI \\
        \midrule
        \multicolumn{11}{c}{\textit{--- task-specific finetuning (ours) ---}}\\
        Teacher BERT-base & 78.9 & 57.9 & 84.2 & 84.6 & 91.4 & 89.7 & 67.9 & 91.7  & 88.0 & 54.9\\
        \midrule
        \multicolumn{11}{c}{\textit{--- task-specific distillation (ours) CBOW not pretrained ---}}\\
        Linear probe & 53.8 & 11.5 & 46.6 & 72.8 & 62.2 & 76.7 & 52.7 & 83.5 & 22.0 & 56.3 \\
        MLP & 61.0 & 14.3 & 57.8 & 77.2 & 70.3 & 86.0 & 56.7 & 82.3 & 47.0 & \textbf{57.7}\\
        CNN & 53.8 & 11.2 & 51.5 & 75.0 & 65.8 & 81.3 & 53.1 & 82.3 & 7.2 & 56.3\\
        BiLSTM & 48.4 & 11.5 & 31.8 & 68.3 & 66.8 & 63.2 & 56.7 & 83.5 & 1.5 & 56.3\\
        \midrule
        \multicolumn{11}{c}{\textit{--- task-specific distillation (ours) CBOW pretrained ---}}\\
        Linear probe & 56.3 & 9.0 & 47.1 & 72.8 & 64.8 & 77.1 & 53.4 & 82.5 & 43.4 & 56.3\\
        MLP & 63.8 & 14.0 & 61.7 & \textbf{78.2} & 70.8 & 86.2 & 57.4 & 83.8 & \textbf{66.0} & 56.3 \\
        CNN & 53.7 & 10.9 & 55.0 & 73.8 & 66.2 & 82.1 & 53.1 & 82.2 & 3.8 & 56.3\\
        BiLSTM & 47.7 & 0 & 32.7 & 68.4 & 69.6 & 63.2 & 55.6 & 82.5 & 1.3 & 56.3\\
        \midrule
        \multicolumn{11}{c}{\textit{--- task-specific distillation (ours) CMOW not pretrained ---}}\\
        Linear probe & 55.1 & 10.9 & 54.3 & 71.8 & 62.7 & 80.9 & 56.0 & 85.2  & 17.6 & 56.3\\
        MLP & 63.2 & 14.2 & 61.9 & 75.5 & 72.4 & 86.3 & 55.2 & 83.7 & 62.7 & 56.3\\
        CNN & 55.4 & 12.4 & 45.3 & 72.3 & 61.5 & 82.6 & 57.4 & 84.3 & 26.1 & 56.3\\
        BiLSTM & 47.5 & 0 & 31.8 & 70.3 & 49.5 & 81.0 & 55.6 & 83.4 & 
        0 & 56.3 \\
        \midrule
        \multicolumn{11}{c}{\textit{--- task-specific distillation (ours) CMOW pretrained ---}}\\
        Linear probe & 56.3 & 22.4 & 48.4 & 72.5 & 61.3 & 81.9 & 54.5 & 83.9 & 24.2 & \textbf{57.7}\\
        MLP & 61.2 & 20.9 & 60.2 & 73.8 & 64.6 & 85.9 & 54.9 & 84.4 & 49.4 & 56.3\\
            CNN & 53.4 & 18.5 & 40.6 & 71.8 & 58.2 & 68.3 & 54.9 & 85.4 & 26.9 & 56.3\\
        BiLSTM & 49.7 & 0 & 32.7 & 68.3 & 67.2 & 82.9 & 57.0 & 82.5 & 0 & 56.3\\
        \midrule
        \multicolumn{11}{c}{\textit{--- task-specific distillation (ours) Hybrid not pretrained ---}}\\
        Linear probe & 51.7 & 11.2 & 39.0 & 71.1 & 49.5 & 81.8 & 56.0 & 85.2 & 14.3 & \textbf{57.7} \\
        MLP & 62.5 & 13.1 & 62.5 & 74.3 & 71.5 & \textbf{86.6} & 58.1 & 83.1 & 58.6 & 56.3\\
        CNN & 52.8 & 11.9 & 45.3 & 71.6 & 61.4 & 84.8 & 55.2 & 85.4 & 2.9 & 56.3\\
        BiLSTM & 50.9 & 0 & 42.6 & 70.1 & 60.3 & 79.3 & 56.0 & 84.4 & 9.3 & 56.3\\
        \midrule
        \multicolumn{11}{c}{\textit{--- task-specific distillation (ours) Hybrid pretrained ---}}\\
        Linear probe & 54.0 & 19.6 & 45.7 & 71.3 & 63.4 & 80.9 & 54.2 & 84.1 & 11.7 & 54.9\\
        MLP & 62.7 & 20.9 & 62.6 & 74.5 & 68.6 & 85.7 & 56.3 & 83.1 & 56.2 & 56.3\\
        CNN & 57.9 & 19.6 & 37.6 & 75.7 & 62.0 & 85.4 & 54.9 & 82.3 & 48.5 & 54.9\\
        BiLSTM & 52.1 & 0 & 48.0 & 68.4 & 71.9 & 85.3 & 56.7 & 82.5 & 0 & 56.3\\
        \bottomrule
    \end{tabular}
\end{table*}

\begin{table*}[p]
    \centering
    \small
    \caption{Scores on the GLUE development set with DiffCat encoding and the bidirectional CMOW/CBOW-Hybrid model}
    \label{tab:results:bidi}
    \begin{tabular}{lcccccccccc}
        \toprule
        Task-Specific Distillation & Score & CoLA & MNLI-m & MRPC & QNLI & QQP & RTE & SST-2 & STS-B & WNLI \\
        \midrule
        \multicolumn{11}{c}{\textit{--- task-specific finetuning (ours) ---}}\\
        Teacher BERT-base & 78.9 & 57.9 & 84.2 & 84.6 & 91.4 & 89.7 & 67.9 & 91.7  & 88.0 & 54.9\\
        \midrule
        \multicolumn{11}{c}{\textit{--- task-specific distillation (ours) Bidirectional Hybrid, not pretrained ---}}\\
        Linear probe & 53.5 & 11.6 & 39.4 & 71.6 & 64.3 & 82.5 & 56.3 & 85.0 & 14.6 & 56.3\\
        MLP & 63.2 & 13.0 & \textbf{63.3} & 75.7 & \textbf{72.6} & 86.1 & 57.4 & 83.3 & 59.7 & 57.7\\
        CNN & 52.7 & 14.5 & 37.3 & 71.3 & 60.8 & 86.4 & 55.2 & 85.8 & 6.6 & 56.3\\
        \midrule
        \multicolumn{11}{c}{\textit{--- task-specific distillation (ours) Bidirectional Hybrid, pretrained ---}}\\
        Linear probe & 55.5 & 18.1 & 42.4 & 72.1 & 64.9 & 81.2 & 56.7 & 85.2 & 22.5 & 56.3\\
        MLP & \textbf{64.6} & \textbf{23.3} & 61.8 & 75.0 & 72.0 & 86.3 & \textbf{59.9} & 82.9 & 62.9 & 57.7 \\
        CNN & 55.1 & 20.5 & 39.3 & 73.8 & 61.3 & 85.9 & 56.3 & 85.5 & 15.9 & 57.7\\
        \bottomrule
    \end{tabular}
\end{table*}

\begin{table*}[p]
    \centering
    \small
    \caption{Number of parameters without DiffCat encoding}
    \label{tab:parameters}
    \begin{tabular}{lrrr}
        \toprule
            & CoLA, MRPC, QNLI, QQP, SST-2, RTE, WNLI & MNLI & STS-B \\
        \midrule
        \multicolumn{3}{c}{\textit{--- task-specific distillation (ours) CBOW ---}}\\
        Linear probe & 47,861,634 & 47,862,419 & 47,876,549 \\
        \textit{-- only classifier} & 3,138 & 3,923 & 18,053 \\
        MLP & 48,647,498 & 48,648,499 & 48,666,517 \\
        \textit{-- only classifier}  & 789,002 & 790,003 & 808,021 \\
        CNN & 47,862,708 & 47,864,737 & 47,901,259 \\
        \textit{-- only classifier}  & 4,212 & 6,241 & 42,763\\
        BiLSTM & 53,704,002 & 53,705,027 & 53,723,477 \\
        \textit{-- only classifier}  & 5,845,506 & 5,846,531 & 5,864,981 \\
        \midrule
        \multicolumn{3}{c}{\textit{--- task-specific distillation (ours) CMOW ---}}\\
        Linear probe & 23,932,386 & 23,933,171 & 23,947,301 \\
        \textit{-- only classifier} & 3,138 & 3,923 & 18,053\\
        MLP & 24,718,250 & 24,719,251 & 24,737,269 \\
        \textit{-- only classifier}  & 789,002 & 790,003 & 808,021 \\
        CNN & 23,933,460 & 23,935,489 & 23,972,011 \\
        \textit{-- only classifier}  & 4,212 & 6,241 & 42,763\\
        BiLSTM & 24,853,978 & 24,854,371 & 35,022,869 \\
        \textit{-- only classifier}  & 924,730 & 925,123 & 110,936,21\\
        \midrule
        \multicolumn{3}{c}{\textit{--- task-specific distillation (ours) Hybrid ---}}\\
        Linear probe & 24,420,802 & 24,421,603 & 24,436,021 \\
        \textit{-- only classifier}  & 3,202 & 4,003 & 18,421\\
        MLP & 25,222,602 & 25,223,603 & 25,241,621  \\
        \textit{-- only classifier}  & 805,002 & 806,003 & 824,021\\
        CNN & 24,420,558 & 24,421,855 & 24,445,201 \\
        \textit{-- only classifier}  & 2,958 & 4,255 & 27,601\\
        BiLSTM & 30,328,642 & 30,329,667 & 30,348,117 \\
        \textit{-- only classifier}  & 5,911,042 & 5,912,067 & 5,930,517 \\
         \midrule
        \bottomrule
    \end{tabular}
    
\end{table*}

\begin{table*}[p]
    \centering
    \small
    \caption{Number of parameters with DiffCat two-sequence encoding}
    \label{tab:parameters:sia}
    \begin{tabular}{lrrr}
        \toprule
            & CoLA, MRPC, QNLI, QQP, SST-2, RTE, WNLI & MNLI & STS-B \\
        \midrule
        \multicolumn{3}{c}{\textit{--- task-specific distillation (ours) CBOW ---}}\\
        Linear probe & 47,867,906 & 47,870,259 & 47,912,613\\
        \textit{-- only classifier} & 9,410 & 11,763 & 54,117 \\
        MLP & 50,215,498 & 50,216,499 & 50,234,517 \\
        \textit{-- only classifier} & 2,357,002 & 2,358,003 & 2,376,021 \\
        CNN & 47,865,932 & 47,869,313 & 47,930,171 \\
        \textit{-- only classifier} & 7,436 & 10,817 & 71,675\\
        BiLSTM & 147,477,458 & 147,479,811 & 147,522,165 \\
        \textit{-- only classifier} & 99,618,962 & 99,621,315 & 99,663,669 \\
        \midrule
        \multicolumn{3}{c}{\textit{--- task-specific distillation (ours) CMOW ---}}\\
        Linear probe & 23,938,658 & 23,941,011 & 23,983,365 \\
        \textit{-- only classifier} & 9,410 & 11,763 & 54,117\\
        MLP & 26,286,250 & 26,287,251 & 26,305,269 \\
        \textit{-- only classifier} & 2,357,002 & 2,358,003 & 2,376,021 \\
        CNN & 23,936,684 & 23,940,065 & 24,000,923 \\
        \textit{-- only classifier} & 7,436 & 10,817 & 71,675\\
        BiLSTM & 123,548,210 & 123,550,563 & 123,592,917 \\
        \textit{-- only classifier} & 99,618,962 & 99,621,315 & 99,663,669 \\
        \midrule
        \multicolumn{3}{c}{\textit{--- task-specific distillation (ours) Hybrid ---}}\\
        Linear probe & 24,427,202 & 24,429,603 & 24,472,821 \\
        \textit{-- only classifier} & 9,602 & 12,003 & 55,221\\
        MLP & 26,822,602 & 26,823,603 & 26,841,621 \\
        \textit{-- only classifier} & 2,405,002 & 2,406,003 & 2,424,021\\
        CNN & 24,424,990 & 24,427,583 & 24,474,257 \\
        \textit{-- only classifier} & 7,390 & 9,983 & 56,657\\
        BiLSTM & 128,143,202 & 128,145,603 & 128,188,821 \\
        \textit{-- only classifier} & 103,725,602 & 103,728,003 & 103,771,221 \\
         \midrule
         \multicolumn{3}{c}{\textit{--- task-specific distillation (ours) Hybrid bidirectional ---}}\\
        Linear probe & 36,640,802 & 36,644,403 & 36,709,221 \\
        \textit{-- only classifier} & 14402 & 18003 & 82,821\\
        MLP & 40,231,402 & 40,232,403 & 4,025,0421 \\
        \textit{-- only classifier} & 3,605,002 & 3,606,003 & 3,624,021 \\
        CNN & 36,638,164 & 36,641,729 & 36,705,899 \\
        \textit{-- only classifier} & 11,764 & 15,329 & 79,499 \\
        BiLSTM & 451,437,602 &  & 451,528,821 \\
        \textit{-- only classifier} & 414,811,202 &  & 414,902,421 \\
        \bottomrule
    \end{tabular}
    
\end{table*}

%% file: main.bbl
\begin{thebibliography}{10}
\providecommand{\url}[1]{#1}
\csname url@samestyle\endcsname
\providecommand{\newblock}{\relax}
\providecommand{\bibinfo}[2]{#2}
\providecommand{\BIBentrySTDinterwordspacing}{\spaceskip=0pt\relax}
\providecommand{\BIBentryALTinterwordstretchfactor}{4}
\providecommand{\BIBentryALTinterwordspacing}{\spaceskip=\fontdimen2\font plus
\BIBentryALTinterwordstretchfactor\fontdimen3\font minus
  \fontdimen4\font\relax}
\providecommand{\BIBforeignlanguage}[2]{{%
\expandafter\ifx\csname l@#1\endcsname\relax
\typeout{** WARNING: IEEEtran.bst: No hyphenation pattern has been}%
\typeout{** loaded for the language `#1'. Using the pattern for}%
\typeout{** the default language instead.}%
\else
\language=\csname l@#1\endcsname
\fi
#2}}
\providecommand{\BIBdecl}{\relax}
\BIBdecl

\bibitem{bert}
\BIBentryALTinterwordspacing
J.~Devlin, M.~Chang, K.~Lee, and K.~Toutanova, ``{BERT:} pre-training of deep
  bidirectional transformers for language understanding,'' in \emph{{NAACL-HLT}
  {(1)}}.\hskip 1em plus 0.5em minus 0.4em\relax Association for Computational
  Linguistics, 2019, pp. 4171--4186. [Online]. Available:
  \url{https://www.aclweb.org/anthology/N19-1423}
\BIBentrySTDinterwordspacing

\bibitem{T5}
\BIBentryALTinterwordspacing
C.~Raffel, N.~Shazeer, A.~Roberts, K.~Lee, S.~Narang, M.~Matena, Y.~Zhou,
  W.~Li, and P.~J. Liu, ``Exploring the limits of transfer learning with a
  unified text-to-text transformer,'' \emph{ArXiv}, vol. abs/1910.10683, 2020.
  [Online]. Available: \url{https://arxiv.org/abs/1910.10683}
\BIBentrySTDinterwordspacing

\bibitem{glue}
\BIBentryALTinterwordspacing
A.~Wang, A.~Singh, J.~Michael, F.~Hill, O.~Levy, and S.~R. Bowman, ``{GLUE:}
  {A} multi-task benchmark and analysis platform for natural language
  understanding,'' in \emph{BlackboxNLP@EMNLP}.\hskip 1em plus 0.5em minus
  0.4em\relax Association for Computational Linguistics, 2018, pp. 353--355.
  [Online]. Available: \url{https://doi.org/10.18653/v1/w18-5446}
\BIBentrySTDinterwordspacing

\bibitem{superglue}
\BIBentryALTinterwordspacing
A.~Wang, Y.~Pruksachatkun, N.~Nangia, A.~Singh, J.~Michael, F.~Hill, O.~Levy,
  and S.~R. Bowman, ``{SuperGLUE}: {A} stickier benchmark for general-purpose
  language understanding systems,'' in \emph{NeurIPS}, 2019, pp. 3261--3275.
  [Online]. Available:
  \url{https://proceedings.neurips.cc/paper/2019/hash/4496bf24afe7fab6f046bf4923da8de6-Abstract.html}
\BIBentrySTDinterwordspacing

\bibitem{GPT-3}
\BIBentryALTinterwordspacing
T.~B. Brown, B.~Mann, N.~Ryder, M.~Subbiah, J.~Kaplan, and {others}, ``Language
  models are few-shot learners,'' \emph{ArXiv}, vol. abs/2005.14165, 2020.
  [Online]. Available: \url{https://arxiv.org/abs/2005.14165}
\BIBentrySTDinterwordspacing

\bibitem{DBLP:conf/acl/StrubellGM19}
\BIBentryALTinterwordspacing
E.~Strubell, A.~Ganesh, and A.~McCallum, ``Energy and policy considerations for
  deep learning in {NLP},'' in \emph{{ACL} {(1)}}.\hskip 1em plus 0.5em minus
  0.4em\relax Association for Computational Linguistics, 2019, pp. 3645--3650.
  [Online]. Available: \url{https://doi.org/10.18653/v1/p19-1355}
\BIBentrySTDinterwordspacing

\bibitem{foundation-models}
\BIBentryALTinterwordspacing
R.~Bommasani, D.~A. Hudson, E.~Adeli, R.~Altman, S.~Arora, and {others}, ``On
  the opportunities and risks of foundation models,'' \emph{ArXiv}, vol.
  abs/2108.07258, 2021. [Online]. Available:
  \url{https://arxiv.org/abs/2108.07258}
\BIBentrySTDinterwordspacing

\bibitem{DBLP:conf/nips/Sanh0R20}
\BIBentryALTinterwordspacing
V.~Sanh, T.~Wolf, and A.~M. Rush, ``Movement pruning: Adaptive sparsity by
  fine-tuning,'' in \emph{NeurIPS}, 2020. [Online]. Available:
  \url{https://proceedings.neurips.cc/paper/2020/hash/eae15aabaa768ae4a5993a8a4f4fa6e4-Abstract.html}
\BIBentrySTDinterwordspacing

\bibitem{knowledgedistillation}
\BIBentryALTinterwordspacing
G.~Hinton, O.~Vinyals, and J.~Dean, ``Distilling the knowledge in a neural
  network,'' \emph{ArXiv}, vol. abs/1503.02531, 2015. [Online]. Available:
  \url{https://arxiv.org/abs/1503.02531}
\BIBentrySTDinterwordspacing

\bibitem{modelcompression}
\BIBentryALTinterwordspacing
C.~Bucila, R.~Caruana, and A.~Niculescu{-}Mizil, ``Model compression,'' in
  \emph{{KDD}}.\hskip 1em plus 0.5em minus 0.4em\relax {ACM}, 2006, pp.
  535--541. [Online]. Available: \url{https://doi.org/10.1145/1150402.1150464}
\BIBentrySTDinterwordspacing

\bibitem{distilbert}
\BIBentryALTinterwordspacing
V.~Sanh, L.~Debut, J.~Chaumond, and T.~Wolf, ``{DistilBERT}, a distilled
  version of bert: smaller, faster, cheaper and lighter,'' \emph{ArXiv}, vol.
  abs/1910.01108, 2020. [Online]. Available:
  \url{https://arxiv.org/abs/1910.01108}
\BIBentrySTDinterwordspacing

\bibitem{tinybert}
\BIBentryALTinterwordspacing
X.~Jiao, Y.~Yin, L.~Shang, X.~Jiang, X.~Chen, L.~Li, F.~Wang, and Q.~Liu,
  ``{TinyBERT}: Distilling {BERT} for natural language understanding,'' in
  \emph{{EMNLP} (Findings)}.\hskip 1em plus 0.5em minus 0.4em\relax Association
  for Computational Linguistics, 2020, pp. 4163--4174. [Online]. Available:
  \url{https://doi.org/10.18653/v1/2020.findings-emnlp.372}
\BIBentrySTDinterwordspacing

\bibitem{sun2020mobilebert}
\BIBentryALTinterwordspacing
Z.~Sun, H.~Yu, X.~Song, R.~Liu, Y.~Yang, and D.~Zhou, ``{MobileBERT}: a compact
  task-agnostic {BERT} for resource-limited devices,'' in \emph{{ACL}}.\hskip
  1em plus 0.5em minus 0.4em\relax Association for Computational Linguistics,
  2020, pp. 2158--2170. [Online]. Available:
  \url{https://doi.org/10.18653/v1/2020.acl-main.195}
\BIBentrySTDinterwordspacing

\bibitem{tang2019distilling}
\BIBentryALTinterwordspacing
R.~Tang, Y.~Lu, L.~Liu, L.~Mou, O.~Vechtomova, and J.~Lin, ``Distilling
  task-specific knowledge from bert into simple neural networks,''
  \emph{ArXiv}, vol. abs/1903.12136, 2019. [Online]. Available:
  \url{https://arxiv.org/abs/1903.12136}
\BIBentrySTDinterwordspacing

\bibitem{boundariesbertdistillation}
\BIBentryALTinterwordspacing
M.~Wasserblat, O.~Pereg, and P.~Izsak, ``Exploring the boundaries of
  low-resource {BERT} distillation,'' in \emph{Proceedings of SustaiNLP:
  Workshop on Simple and Efficient Natural Language Processing}.\hskip 1em plus
  0.5em minus 0.4em\relax Association for Computational Linguistics, Nov. 2020,
  pp. 35--40. [Online]. Available:
  \url{https://aclanthology.org/2020.sustainlp-1.5}
\BIBentrySTDinterwordspacing

\bibitem{DBLP:journals/neco/HochreiterS97}
\BIBentryALTinterwordspacing
S.~Hochreiter and J.~Schmidhuber, ``Long short-term memory,'' \emph{Neural
  Comput.}, vol.~9, no.~8, pp. 1735--1780, 1997. [Online]. Available:
  \url{https://doi.org/10.1162/neco.1997.9.8.1735}
\BIBentrySTDinterwordspacing

\bibitem{DBLP:conf/icml/CollobertW08}
\BIBentryALTinterwordspacing
R.~Collobert and J.~Weston, ``A unified architecture for natural language
  processing: deep neural networks with multitask learning,'' in
  \emph{{ICML}}.\hskip 1em plus 0.5em minus 0.4em\relax {ACM}, 2008, pp.
  160--167. [Online]. Available: \url{https://doi.org/10.1145/1390156.1390177}
\BIBentrySTDinterwordspacing

\bibitem{word2vec}
\BIBentryALTinterwordspacing
T.~Mikolov, K.~Chen, G.~Corrado, and J.~Dean, ``Efficient estimation of word
  representations in vector space,'' \emph{ArXiv}, vol. abs/1301.3781, 2013.
  [Online]. Available: \url{https://arxiv.org/abs/1301.3781}
\BIBentrySTDinterwordspacing

\bibitem{cmow}
\BIBentryALTinterwordspacing
F.~Mai, L.~Galke, and A.~Scherp, ``{CBOW} is not all you need: Combining {CBOW}
  with the compositional matrix space model,'' in \emph{{ICLR} (Poster)}.\hskip
  1em plus 0.5em minus 0.4em\relax OpenReview.net, 2019. [Online]. Available:
  \url{https://openreview.net/forum?id=H1MgjoR9tQ}
\BIBentrySTDinterwordspacing

\bibitem{DBLP:conf/acl/RudolphG10}
\BIBentryALTinterwordspacing
S.~Rudolph and E.~Giesbrecht, ``Compositional matrix-space models of
  language,'' in \emph{{ACL}}.\hskip 1em plus 0.5em minus 0.4em\relax The
  Association for Computer Linguistics, 2010, pp. 907--916. [Online].
  Available: \url{https://aclanthology.org/P10-1093/}
\BIBentrySTDinterwordspacing

\bibitem{DBLP:conf/naacl/PetersNIGCLZ18}
\BIBentryALTinterwordspacing
M.~E. Peters, M.~Neumann, M.~Iyyer, M.~Gardner, C.~Clark, K.~Lee, and
  L.~Zettlemoyer, ``Deep contextualized word representations,'' in
  \emph{{NAACL-HLT}}.\hskip 1em plus 0.5em minus 0.4em\relax Association for
  Computational Linguistics, 2018, pp. 2227--2237. [Online]. Available:
  \url{https://doi.org/10.18653/v1/n18-1202}
\BIBentrySTDinterwordspacing

\bibitem{DBLP:journals/tsp/SchusterP97}
\BIBentryALTinterwordspacing
M.~Schuster and K.~K. Paliwal, ``Bidirectional recurrent neural networks,''
  \emph{{IEEE} Trans. Signal Process.}, vol.~45, no.~11, pp. 2673--2681, 1997.
  [Online]. Available: \url{https://doi.org/10.1109/78.650093}
\BIBentrySTDinterwordspacing

\bibitem{DBLP:conf/acl/MouMLX0YJ16}
\BIBentryALTinterwordspacing
L.~Mou, R.~Men, G.~Li, Y.~Xu, L.~Zhang, R.~Yan, and Z.~Jin, ``Natural language
  inference by tree-based convolution and heuristic matching,'' in \emph{{ACL}
  {(2)}}.\hskip 1em plus 0.5em minus 0.4em\relax The Association for Computer
  Linguistics, 2016. [Online]. Available:
  \url{https://aclanthology.org/P16-2022/}
\BIBentrySTDinterwordspacing

\bibitem{sentenceBert}
\BIBentryALTinterwordspacing
N.~Reimers and I.~Gurevych, ``{Sentence-BERT}: {Sentence Embeddings using
  Siamese BERT-Networks},'' in \emph{{EMNLP/IJCNLP} {(1)}}.\hskip 1em plus
  0.5em minus 0.4em\relax Association for Computational Linguistics, 2019, pp.
  3980--3990. [Online]. Available: \url{https://doi.org/10.18653/v1/D19-1410}
\BIBentrySTDinterwordspacing

\bibitem{roberta2019}
\BIBentryALTinterwordspacing
Y.~Liu, M.~Ott, N.~Goyal, J.~Du, M.~Joshi, D.~Chen, O.~Levy, M.~Lewis,
  L.~Zettlemoyer, and V.~Stoyanov, ``{RoBERTa}: {A} robustly optimized {BERT}
  pretraining approach,'' \emph{CoRR}, vol. abs/1907.11692, 2019. [Online].
  Available: \url{https://arxiv.org/abs/1907.11692}
\BIBentrySTDinterwordspacing

\bibitem{DBLP:conf/iccv/ZhuKZSUTF15}
\BIBentryALTinterwordspacing
Y.~Zhu, R.~Kiros, R.~S. Zemel, R.~Salakhutdinov, R.~Urtasun, A.~Torralba, and
  S.~Fidler, ``Aligning books and movies: Towards story-like visual
  explanations by watching movies and reading books,'' in \emph{{ICCV}}.\hskip
  1em plus 0.5em minus 0.4em\relax {IEEE} Computer Society, 2015, pp. 19--27.
  [Online]. Available: \url{https://ieeexplore.ieee.org/document/7410368}
\BIBentrySTDinterwordspacing

\bibitem{wordpiece}
\BIBentryALTinterwordspacing
Y.~Wu, M.~Schuster, Z.~Chen, Q.~V. Le, M.~Norouzi, and {others}, ``Google's
  neural machine translation system: Bridging the gap between human and machine
  translation,'' \emph{ArXiv}, vol. abs/1609.08144, 2016. [Online]. Available:
  \url{https://arxiv.org/abs/1609.08144}
\BIBentrySTDinterwordspacing

\bibitem{phua2021word2rate}
\BIBentryALTinterwordspacing
G.~Phua, S.~Lin, and D.~Poletti, ``Word2rate: training and evaluating multiple
  word embeddings as statistical transitions,'' \emph{ArXiv}, vol.
  abs/2104.08173, 2021. [Online]. Available:
  \url{https://arxiv.org/abs/2104.08173}
\BIBentrySTDinterwordspacing

\bibitem{sun2019patient}
\BIBentryALTinterwordspacing
S.~Sun, Y.~Cheng, Z.~Gan, and J.~Liu, ``Patient knowledge distillation for
  {BERT} model compression,'' in \emph{{EMNLP/IJCNLP} {(1)}}.\hskip 1em plus
  0.5em minus 0.4em\relax Association for Computational Linguistics, 2019, pp.
  4322--4331. [Online]. Available: \url{https://doi.org/10.18653/v1/D19-1441}
\BIBentrySTDinterwordspacing

\bibitem{ladabert}
\BIBentryALTinterwordspacing
Y.~Mao, Y.~Wang, C.~Wu, C.~Zhang, Y.~Wang, Q.~Zhang, Y.~Yang, Y.~Tong, and
  J.~Bai, ``Ladabert: Lightweight adaptation of {BERT} through hybrid model
  compression,'' in \emph{{COLING}}.\hskip 1em plus 0.5em minus 0.4em\relax
  International Committee on Computational Linguistics, 2020, pp. 3225--3234.
  [Online]. Available: \url{https://doi.org/10.18653/v1/2020.coling-main.287}
\BIBentrySTDinterwordspacing

\bibitem{DBLP:conf/emnlp/TsaiRJALA19}
\BIBentryALTinterwordspacing
H.~Tsai, J.~Riesa, M.~Johnson, N.~Arivazhagan, X.~Li, and A.~Archer, ``Small
  and practical {BERT} models for sequence labeling,'' in \emph{{EMNLP/IJCNLP}
  {(1)}}.\hskip 1em plus 0.5em minus 0.4em\relax Association for Computational
  Linguistics, 2019, pp. 3630--3634. [Online]. Available:
  \url{https://aclanthology.org/D19-1374/}
\BIBentrySTDinterwordspacing

\bibitem{yang2019multitaskKD}
\BIBentryALTinterwordspacing
Z.~Yang, L.~Shou, M.~Gong, W.~Lin, and D.~Jiang, ``Model compression with
  multi-task knowledge distillation for web-scale question answering system,''
  \emph{Arxiv}, vol. abs/1904.09636, 2019. [Online]. Available:
  \url{https://arxiv.org/abs/1904.09636}
\BIBentrySTDinterwordspacing

\bibitem{turc2019wellread}
\BIBentryALTinterwordspacing
I.~Turc, M.~Chang, K.~Lee, and K.~Toutanova, ``Well-read students learn better:
  The impact of student initialization on knowledge distillation,''
  \emph{Arxiv}, vol. abs/1908.08962, 2019. [Online]. Available:
  \url{https://arxiv.org/abs/1908.08962}
\BIBentrySTDinterwordspacing

\bibitem{DBLP:conf/icml/GuptaAGN15}
\BIBentryALTinterwordspacing
S.~Gupta, A.~Agrawal, K.~Gopalakrishnan, and P.~Narayanan, ``Deep learning with
  limited numerical precision,'' in \emph{{ICML}}, ser. {JMLR} Workshop and
  Conference Proceedings, vol.~37.\hskip 1em plus 0.5em minus 0.4em\relax
  JMLR.org, 2015, pp. 1737--1746. [Online]. Available:
  \url{http://proceedings.mlr.press/v37/gupta15.html}
\BIBentrySTDinterwordspacing

\bibitem{wu2020integer}
\BIBentryALTinterwordspacing
H.~Wu, P.~Judd, X.~Zhang, M.~Isaev, and P.~Micikevicius, ``Integer quantization
  for deep learning inference: Principles and empirical evaluation,''
  \emph{ArXiv}, vol. abs/2004.09602, 2020. [Online]. Available:
  \url{https://arxiv.org/abs/2004.09602}
\BIBentrySTDinterwordspacing

\bibitem{efficient-transformers}
\BIBentryALTinterwordspacing
Y.~Tay, M.~Dehghani, D.~Bahri, and D.~Metzler, ``Efficient transformers: A
  survey,'' \emph{ArXiv}, vol. abs/2009.06732, 2020. [Online]. Available:
  \url{https://arxiv.org/abs/2009.06732}
\BIBentrySTDinterwordspacing

\bibitem{DBLP:conf/iclr/WietingK19}
\BIBentryALTinterwordspacing
J.~Wieting and D.~Kiela, ``No training required: Exploring random encoders for
  sentence classification,'' in \emph{{ICLR} (Poster)}.\hskip 1em plus 0.5em
  minus 0.4em\relax OpenReview.net, 2019. [Online]. Available:
  \url{https://openreview.net/forum?id=BkgPajAcY7}
\BIBentrySTDinterwordspacing

\bibitem{liu2021pay}
\BIBentryALTinterwordspacing
H.~Liu, Z.~Dai, D.~R. So, and Q.~V. Le, ``Pay attention to {MLP}s,''
  \emph{ArXiv}, vol. abs/2105.08050, 2021. [Online]. Available:
  \url{https://arxiv.org/abs/2105.08050}
\BIBentrySTDinterwordspacing

\bibitem{DBLP:conf/fat/BenderGMS21}
\BIBentryALTinterwordspacing
E.~M. Bender, T.~Gebru, A.~McMillan{-}Major, and S.~Shmitchell, ``On the
  dangers of stochastic parrots: Can language models be too big?'' in
  \emph{FAccT}.\hskip 1em plus 0.5em minus 0.4em\relax {ACM}, 2021, pp.
  610--623. [Online]. Available: \url{https://doi.org/10.1145/3442188.3445922}
\BIBentrySTDinterwordspacing

\bibitem{hooker2020characterising}
\BIBentryALTinterwordspacing
S.~Hooker, N.~Moorosi, G.~Clark, S.~Bengio, and E.~Denton, ``Characterising
  bias in compressed models,'' \emph{Arxiv}, vol. abs/2010.03058, 2020.
  [Online]. Available: \url{https://arxiv.org/abs/2010.03058}
\BIBentrySTDinterwordspacing

\bibitem{DBLP:conf/blackboxnlp/SoulosMLS20}
\BIBentryALTinterwordspacing
P.~Soulos, R.~T. McCoy, T.~Linzen, and P.~Smolensky, ``Discovering the
  compositional structure of vector representations with role learning
  networks,'' in \emph{BlackboxNLP@EMNLP}.\hskip 1em plus 0.5em minus
  0.4em\relax Association for Computational Linguistics, 2020, pp. 238--254.
  [Online]. Available: \url{https://doi.org/10.18653/v1/2020.blackboxnlp-1.23}
\BIBentrySTDinterwordspacing

\bibitem{sent_analysis_bilstm}
\BIBentryALTinterwordspacing
G.~{Xu}, Y.~{Meng}, X.~{Qiu}, Z.~{Yu}, and X.~{Wu}, ``Sentiment analysis of
  comment texts based on bilstm,'' \emph{IEEE Access}, vol.~7, pp.
  51\,522--51\,532, 2019. [Online]. Available:
  \url{https://doi.org/10.1109/ACCESS.2019.2909919}
\BIBentrySTDinterwordspacing

\bibitem{single_bilstm}
\BIBentryALTinterwordspacing
Z.~{Hameed} and B.~{Garcia-Zapirain}, ``Sentiment classification using a
  single-layered bilstm model,'' \emph{IEEE Access}, vol.~8, pp.
  73\,992--74\,001, 2020. [Online]. Available:
  \url{https://doi.org/10.1109/ACCESS.2020.2988550}
\BIBentrySTDinterwordspacing

\bibitem{mishra2017apprentice}
\BIBentryALTinterwordspacing
A.~K. Mishra and D.~Marr, ``Apprentice: Using knowledge distillation techniques
  to improve low-precision network accuracy,'' in \emph{{ICLR} (Poster)}.\hskip
  1em plus 0.5em minus 0.4em\relax OpenReview.net, 2018. [Online]. Available:
  \url{https://openreview.net/forum?id=B1ae1lZRb}
\BIBentrySTDinterwordspacing

\bibitem{polino2018model}
\BIBentryALTinterwordspacing
A.~Polino, R.~Pascanu, and D.~Alistarh, ``Model compression via distillation
  and quantization,'' in \emph{{ICLR} (Poster)}.\hskip 1em plus 0.5em minus
  0.4em\relax OpenReview.net, 2018. [Online]. Available:
  \url{https://openreview.net/forum?id=S1XolQbRW}
\BIBentrySTDinterwordspacing

\bibitem{chen2017learning}
\BIBentryALTinterwordspacing
G.~Chen, W.~Choi, X.~Yu, T.~X. Han, and M.~Chandraker, ``Learning efficient
  object detection models with knowledge distillation,'' in \emph{{NIPS}},
  2017, pp. 742--751. [Online]. Available:
  \url{https://proceedings.neurips.cc/paper/2017/hash/e1e32e235eee1f970470a3a6658dfdd5-Abstract.html}
\BIBentrySTDinterwordspacing

\bibitem{mukherjee2020distilling}
\BIBentryALTinterwordspacing
S.~Mukherjee and A.~H. Awadallah, ``Distilling bert into simple neural networks
  with unlabeled transfer data,'' \emph{ArXiv}, vol. abs/1910.01769, 2020.
  [Online]. Available: \url{https://arxiv.org/abs/1910.01769}
\BIBentrySTDinterwordspacing

\end{thebibliography}
